\documentclass[sigconf]{acmart}

\AtBeginDocument{%
  \providecommand\BibTeX{{%
    \normalfont B\kern-0.5em{\scshape i\kern-0.25em b}\kern-0.8em\TeX}}}

\copyrightyear{2021}
\acmYear{2021}
\setcopyright{acmcopyright}\acmConference[ICMI '21]{Proceedings of the 2021 International Conference on Multimodal Interaction}{October 18--22, 2021}{Montréal, QC, Canada}
\acmBooktitle{Proceedings of the 2021 International Conference on Multimodal Interaction (ICMI '21), October 18--22, 2021, Montréal, QC, Canada}
\acmPrice{15.00}
\acmDOI{10.1145/3462244.3479932}
\acmISBN{978-1-4503-8481-0/21/10}

\begin{document}

%%
%% The "title" command has an optional parameter,
%% allowing the author to define a "short title" to be used in page headers.
\title{Toddler-Guidance Learning: Impacts of Critical Period on Multimodal AI Agents}
%%
%% The "author" command and its associated commands are used to define
%% the authors and their affiliations.
%% Of note is the shared affiliation of the first two authors, and the
%% "authornote" and "authornotemark" commands
%% used to denote shared contribution to the research.
\author{Junseok Park}
%%\authornote{Both authors contributed equally to this research.}
%%\email{trovato@corporation.com}
%%\orcid{1234-5678-9012}
%%\author{G.K.M. Tobin}
%%\authornotemark[1]
%%\email{webmaster@marysville-ohio.com}
\affiliation{%
  \institution{Seoul National University}
  \streetaddress{Gwanak-ro}
  \city{Seoul}
 %% \state{Ohio}
  \country{Republic of Korea}
  %%\postcode{43017-6221}
}

\author{Kwanyoung Park}
\affiliation{%
  \institution{Seoul National University}
  \streetaddress{Gwanak-ro}
  \city{Seoul}
 %\state{Ohio}
  \country{Republic of Korea}}
 %%\email{larst@affiliation.org}

\author{Hyunseok Oh}
\affiliation{%
 \institution{Seoul National University}
 \streetaddress{Gwanak-ro}
  \city{Seoul}
 %\state{Ohio}
  \country{Republic of Korea}}
 %%\email{larst@affiliation.org}

\author{Ganghun Lee}
\affiliation{%
 \institution{Seoul National University}
 \streetaddress{Gwanak-ro}
  \city{Seoul}
 %\state{Ohio}
  \country{Republic of Korea}}
 %%\email{larst@affiliation.org}

\author{Minsu Lee}
\affiliation{%
 \institution{AIIS, Seoul National University}
 \streetaddress{Gwanak-ro}
  \city{Seoul}
 %\state{Ohio}
  \country{Republic of Korea}}
 %%\email{larst@affiliation.org}

\author{Youngki Lee}
\affiliation{%
 \institution{Seoul National University}
 \streetaddress{Gwanak-ro}
  \city{Seoul}
 %\state{Ohio}
  \country{Republic of Korea}}
 %%\email{larst@affiliation.org}

\author{Byoung-Tak Zhang}
\affiliation{%
 \institution{AIIS, Seoul National University}
 \streetaddress{Gwanak-ro}
  \city{Seoul}
 %\state{Ohio}
  \country{Republic of Korea}}
 %%\email{larst@affiliation.org}

%%
%% By default, the full list of authors will be used in the page
%% headers. Often, this list is too long, and will overlap
%% other information printed in the page headers. This command allows
%% the author to define a more concise list
%% of authors' names for this purpose.
%%\renewcommand{\shortauthors}{Trovato and Tobin, et al.}

%%
%% The abstract is a short summary of the work to be presented in the
%% article.
\begin{abstract}
   Critical periods are phases during which a toddler’s brain develops in spurts. To promote children’s cognitive development, proper guidance is critical in this stage.
  However, it is not clear whether such a critical period also exists for the training of AI agents. %, which is inspired by human-learning mechanisms.
  Similar to human toddlers, well-timed guidance and multimodal interactions might significantly enhance the training efficiency of AI agents as well. To validate this hypothesis, we adapt this notion of critical periods to learning in AI agents and investigate the critical period in the virtual environment for AI agents. We formalize the critical period and Toddler-guidance learning in the reinforcement learning (RL) framework. Then, we built up a toddler-like environment with VECA toolkit to mimic human toddlers' learning characteristics. We study three discrete levels of mutual interaction: weak-mentor guidance (sparse reward), moderate mentor guidance (helper-reward), and mentor demonstration (behavioral cloning). We also introduce the EAVE dataset consisting of 30,000 real-world images to fully reflect the toddler's viewpoint. 
  We evaluate the impact of critical periods on AI agents from two perspectives: how and when they are guided best in both uni- and multimodal learning. 
  Our experimental results show that both uni- and multimodal agents with moderate mentor guidance and critical period on 1 million 
  %(uni/multi, 0.03/0.050 average rewards ) 
  and 2 million %(uni/multi, 0.00/0.046 average rewards ) 
  training steps show a noticeable improvement. 
  We validate these results with transfer learning on the EAVE dataset and find the performance advancement on the same critical period and the guidance.%, with 2M-helper (uni/multi, 64.6/65.8\ accuracy) and 1M-helper (uni/multi, 64.0/64.9\ accuracy) models performing best.
\end{abstract}
%%
%% The code below is generated by the tool at http://dl.acm.org/ccs.cfm.
%% Please copy and paste the code instead of the example below.
%%


%%
%% Keywords. The author(s) should pick words that accurately describe
%% the work being presented. Separate the keywords with commas.

%% A "teaser" image appears between the author and affiliation
%% information and the body of the document, and typically spans the
%% page.
\maketitle

\begin{figure}[t!]
    \centering
    \includegraphics[width=0.35\textwidth]{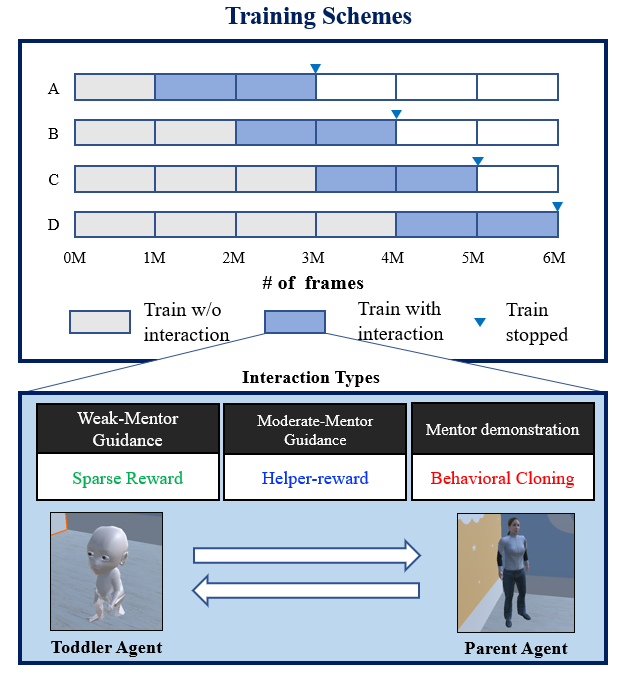}
    \caption{Overall approach to testing the critical period. We first trained the agent for 1/2/3/4 million frames only with sparse rewards. We then continued training the agent with guidance (helper-reward, behavior cloning) or without guidance (sparse reward) for an additional 2 million frames. }
    \label{figure:overallapproach}
    \vspace{-13px}
\end{figure}

%%
%% This command processes the author and affiliation and title
%% information and builds the first part of the formatted document.

\section{Introduction}

Mentor guidance is critical in the early stages of cognitive development. 
Learning in the child stage significantly influences learning-to-learn capabilities like goal-setting and self-control ~\cite{zosh2017learnthruplayreview}. 
%Life experiences and prior knowledge taught in childhood are known to influence the learning in grown adults as well.
For this reason, parent-child interaction is crucial in the overall cognitive development process, including linguistics and object understanding~\cite{olson1990early}. 
%Guided learning for machine agent is thoroughly investigated as a supervision. explicitly supervise through reward function, sensory cues that indirectly associates rewards, or the demonstration which the agent is expected to imitate.
Adequate supervision in the early stages of learning is also important for machine learning (ML), as the sample efficiency of supervised learning
%and semi-supervised learning 
is in general much higher than that of unsupervised approaches~\cite{lu2009semisupervisedlearning}.
%Weakly-Supervised Reinforcement Learning for Controllable Behavior, 
Prior works have investigated diverse forms of guided learning on ML agents such as distillation~\cite{arulkumaran2017briefRLSurvey}, imitation learning, weakly-supervised RL~\cite{lee2020weaklysupervised}. However, there has been limited cross-domain research to find an optimal guidance strategy for a given task. %However, the cross-domain research which controls the level of guidance to find an optimal guidance strategy for a given task does not exist to our knowledge.

In this work, we are inspired by the concept of \emph{critical periods} in the infant-toddler learning, a specific period in which the acquisition of language and visual/auditory processing capabilities are highly accelerated ~\cite{hensch2004critical, robson2002critical, kral2013auditory,  singleton1995agesecondlanguageacquisition,snow1978criticalperiodlang,krashen1973lateralizationcritical,chiswick2008testcriticalperiod}, and
%for development of cognitive tasks such as language acquisition and visual and auditory processing
%It is not only based on the biological development, but also the duration of the parent-child interaction. The separation of mother and child for a certain period after birth causes destructive results in child development. 
we explore the question, "does such a critical period also exist for machine learning (ML) agents?". In particular, we aim to identify the optimal guidance during training on a certain task, and study the correspondence between the optimal duration of guidance in ML algorithms and the critical period in human learning. This study is significant in that the optimal guidance strategy for a task can significantly enhance the training efficiency of the ML model.

%The difference of the real-world toddler and the machine agent in the virtual environment is that the level and the duration of the guidence is easily adjustable. A virtual agent could be reset and trained from the scratch, so the effect of training approaches could be validated in a diverse settings. With parametrization, We can even search the optimal guidence level and the duration through hyperparameter optimization algorithms. We take the advantage of this to discover the effect of different guidence level and duration, and validate the hypothesis that the critical period of guidance exists in machine learning algorithms.

%We define the counterpart of the critical period on an AI agent as an optimal duration of mentor guidance. 
To this end, we define and empirically study the optimal initial time of mentor guidance as the counterpart of the critical period in AI agents.
%We formalize the notions of mentor guidance, optimal guidance policy and the duration of guidance, and analyze ML tasks under our model. 
To do this we formulate a reward structure that is variable through the training iterations, in terms of the Partially Observable Markov Decision Process(POMDP) in RL framework. We then formalize mentor guidance as a policy-invariant mutable reward structure on a given task. Finally, we formally define the optimal initial time of mentor guidance, and study it as the AI agent's equivalent to the \emph{critical period} seen in toddlers.

% We formalize the notions of mentor guidance, optimal guidance policy and the duration of guidance, and analyze ML tasks under our model. We formulate social interaction as well as parent-child interaction in terms of the RL framework. We then define mentor guidance as the 2-agent direct parent-child interaction on a given task. Finally, we define the optimal guidance policy and the duration of guidance given a set of guidances and a task, and match the optimal duration of guidance to the \emph{critical period}.

\begin{figure}[t!]
    \centering
    \includegraphics[width=0.5\textwidth]{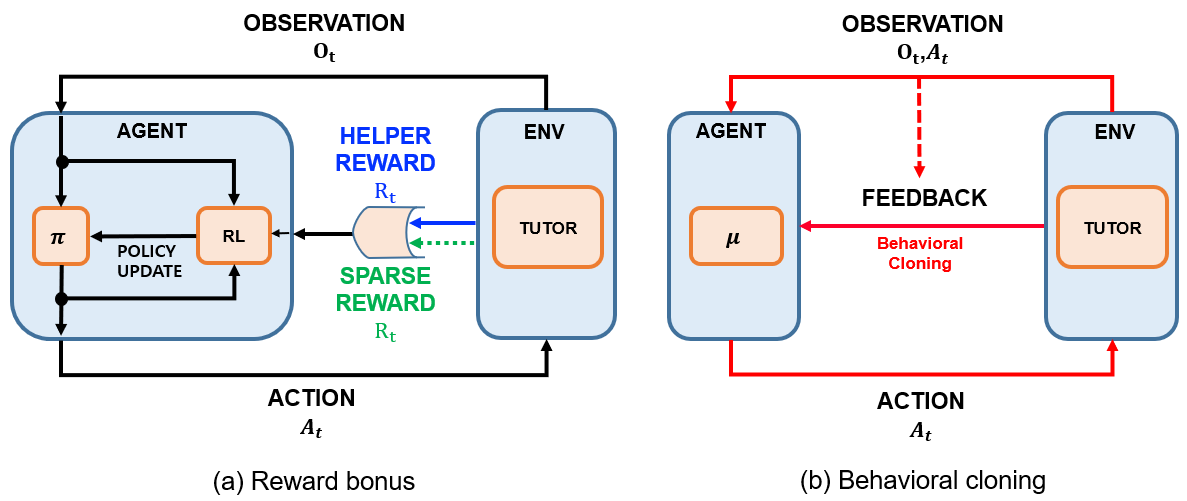}
    \caption{Three types of guidance. (a) Weak-Mentor Guidance (green) and Moderate-Mentor Guidance (blue) use a reward bonus method based on RL. (b) Mentor demonstration (red) use behavioral cloning (Imitation Learning)}
    \label{figure:guidance}
    
     \vspace{-13px}
\end{figure}

Our key finding from experiments is that reinforcement learning agents indeed have a critical period much like toddlers in the moderate guidance learning setting. To demonstrate this, we have implemented guidance inside the VECA~\cite{VECA} virtual environment in a multimodal agent RL framework to model mentor guidance. First, we have modeled learning-through-play of toddlers with an object navigation RL task using VECA. Second, we introduce the EAVE dataset, a real-world image and audio dataset accurately reflecting the toddler's point of view. For our experiments, we have imposed three levels of guidance as seen in Fig~\ref{figure:guidance}: weak-mentor guidance (sparse reward), moderate mentor guidance (helper-reward), and mentor demonstration (behavioral cloning) and we adjust these guidances and their duration to find the best-performing setting. Our studies on unimodal and multi-modal RL agents show that the critical period has appeared only when the toddler agent has moderate guidance while learning. Moreover, we validate our results using the EAVE dataset via transfer learning. Results  confirm that the models trained on their appropriate critical period (1M\&2M iterations) noticeably outperform the models trained after (3M\&4M iterations) their critical period.

%We observed that critical period does exists on ML agents with moderate guidance. We have implemented guidance inside the VECA~\cite{VECA} virtual environment as a multi-agent RL framework to model the mentor guidance. First, we have modeled learning-through-play of toddlers with an object navigation RL task using the VECA. Second, we introduce the EAVE dataset, the real-world image and audio dataset reflecting the toddler's point of view. In studies, we have imposed three levels of guidance: no guidance, moderate mentor guidance, and mentor demonstration and adjusts these guidances and their durations to find the best performing setting. The study1 result shows that the critical period has appeared only when the toddler has moderate guidance learning. Moreover, we validate our results using the EAVE dataset via transfer learning and confirm that, similar to study1, the models with a critical period trained for 2M (64.6\%) and 1M (64\%) frames significantly outperform the ones without a critical period (3M\&4M).
To summarize, this paper's contributions are threefold:
\begin{itemize}
    \item We use a policy-invariant mutable reward structure to formulate the human toddler's critical period in an ML model, especially the multimodal humanoid RL agent we developed.
    \item We construct a multimodal Navigation RL task and a real-world EAVE dataset that can effectively demonstrate the critical period in a multimodal RL agent.
    \item We empirically study the critical period in a unimodal and multimodal RL task with four different periods: 1/2/3/4 million training iterations. Agent trained in a critical period outperforms others significantly with proper guidance. Furthermore, the performance advantage transfers well to the real-world downstream task.
\end{itemize}

\section {Related Works}
Curriculum Learning ~\cite{Bengio2009CurriculumLearning,chen2015weblyCL,shi2015recurrentCL,spitkovsky2009babystepsCL,pentina2015curriculumCL,tudor2016hardCL} is an easy-to-hard training strategy in ML that gradually increases the complexity of the data samples used during the training process~\cite{Soviany2021CurriculumLearningSurvey}. Humans learn the easy and preliminary concepts sooner and the difficult and advanced concepts later. Moreover, humans learn much better when the examples are not randomly presented but are organized in a meaningful order~\cite{Soviany2021CurriculumLearningSurvey,Elman1993Learningsmall}.
Curriculum learning applies a similar strategy for ML model training, and achieves two major advantages: faster convergence and better performance. There is a number of domains including face recognition~\cite{Lin2018ActiveSL}, object segmentation~\cite{Zhang2017CurriculumDAseg} and reinforcement learning~\cite{sutton2018reinforcementlearning} in which curriculum learning has been successfully applied. However, using curriculum learning does not always benefit performance. It requires careful design of the difficulty hierarchy before training. For instance, the diversity of the data samples may be restricted if the task difficulty measure prefers choosing easy examples only from a small subset of classes~\cite{Soviany2020ImageDC,Soviany2021CurriculumLearningSurvey}. An approach diametrical to easy-to-hard curriculum learning, that is emphasizing harder examples as in hard example mining~\cite{Shrivastava2016TrainingRO,Jesson2017CASEDCA} or
anti-curriculum learning~\cite{Pi2016SelfPacedBL,Braun2017ACL}, can achieve improved results in some cases.

ML researchers have recently taken interest in ideas from child learning further technical advances~\cite{bambach2018toddlerinspiredvisualobject,schank1972nlp,turing2009child,tsutsui2020computationalToddler,turchetta2020CISR}.
Learning properly in a child stage is crucial for learning-to-learn capabilities like goal-setting and self-control ~\cite{zosh2017learnthruplayreview,thrun2012learningtolearn}. 
Life experiences and prior knowledge learned in one's childhood are known to influence adult learning as well.
The latest works in deep learning like visual object learning~\cite{bambach2018toddlerinspiredvisualobject} 
are inspired by the way children learn. 
~\citet{achille2018criticalperiodDNN} is the first work to show that deep artificial neural networks also exhibit \emph{critical periods} like humans, a decisive time window for the post-natal developmental stage. In deep neural network training, a rapid
growth in information is followed by a reduction of information from an analysis with Fisher Information. Unfortunately, their analysis is limited to supervised learning in convolution neural networks (CNNs), and not executed on either RL or multimodal setup. To the best of our knowledge, our work is the first attempt to apply the toddler's developmental characteristic of humans to effective RL training. We aim to understand how the functionality and learning-to-learn capabilities are nurtured through appropriate supervision.

Several works in developmental psychology note that more guidance does not mean better human development.
A general understanding of objects develops in an early stage of toddler learning with weak supervision as well as through interactions like mouthing, chewing, and rotating~\cite{gibson1988toddlerlearning,piaget1952origins}.
Furthermore, teaching formal subjects too early for a child is counterproductive in the long run~\cite{suggate2013childlatelearning}. Inspired by human toddler learning, we conjecture that there is also a form of optimal supervision and an optimal time period for RL training of intelligent machines. We empirically search for a type of guidance that accelerates the training, but does not change the optimal policy of the environment.

\section{Method}

% TODO 1/4 page
\paragraph{Reinforcement Learning} 
Reinforcement learning (RL) is a framework for experience-driven autonomous learning~\cite{sutton2018reinforcementlearning} in which an agent interacts with the environment to learn optimal behaviour, improving over
time through trial and error~\cite{arulkumaran2017briefRLSurvey}. Upon observing the consequent state, an agent learns to alter its policy in response to received rewards.
The environment in RL is typically assumed to be a Markov Decision Process (MDP). MDP is memoryless; the conditional probability of the next state and reward are stochastically determined only from the current state and action, and do not depend on the past~\cite{shoham2003MultiagentRLCriticalSurvey}.
\subsection{Guidance in a POMDP}
%We introduce the Multi-agent POMDP to formalize the social interaction from it. %upon the process.

In social psychology, a \emph{social behavior} or \emph{social action} is a behavior that affects the other person and provokes a response~\cite{sztompka2002socjologia,homans1974social}. A general social interaction sequence is defined as a sequence of a person acting toward another (under the expectation), and the other acting towards the person in response (under the interpretation)~\cite{darley1980expectancysocialinteraction}. Guided learning, or \emph{mentor guidance}, can be defined as a process in which learners achieve their learning goal with the supervision of a more experienced mentor~\cite{billett2000guided}. % or indirect media like text or videos.

Inspired by these psychological discoveries, we formalize the notion of \emph{Guidance} in the context of Partially Observable Markov Decision Process (POMDP). We find it analogous of a person as an agent, and an effect of the behavior as a change in the observation.

Partially-observable Markov decision processes (POMDP) extend RL frameworks so that an agent observes only part of the world state~\cite{doshi2009infinitePOMDP}.
%Multi-agent POMDP %https://www.aaai.org/ocs/index.php/AAAI/AAAI15/paper/viewFile/9889/9495extends the formulation of POMDP to multiple agents so that each agent observes only partial states. 
A POMDP is a 7-tuple $\langle S, A, T, R, \Omega, O, \gamma \rangle$ consisting of:  $S$, a set of possible states; $A$, a set of actions; $T$, a set of transition probabilities; $R$, a reward function $R(s, a)$ for a state $s$ and action $a$; $\Omega$, a set of possible observations, $O$, a set of observation probabilities $o(s,a) = Pr(o | a, s)$ for given state $s$ and action $a$; and $\gamma$ a discount factor. We extend this to include a no-op action (do nothing) in the action space as $A' = A \bigcup \{a_{nop}\}$.
% NO-OP action https://www.aaai.org/ocs/index.php/AAAI/AAAI16/paper/download/12389/11847

%이 파트가 위로 올라가는게 나을 거 같습니다.
% 아 넵 감사합니다 넵 정말 감사합니다!!!
% 관련레퍼 찾아볼게요 ! 아래 내용에 관해서 
\subsubsection{Mutable Reward Structure}
%We bring the basis of definition from the Sztompka's work.
%Sztompka, Piotr. 2002. Socjologia, Znak. ISBN 83-240-0218-9. p. 107.
Mutable reward structure is a reward function that is variable under number of iterations. Since MDP is memoryless, the state should incorporate the number of training iteration as a reference. Suppose the state consists of $(s, t)$ where $t \in \mathbb{N}$ is the training iteration. The reward structure $\tilde{R}$ is mutable if such state $s \in S$,  action $a \in A$ exists.
\begin{align}
    \tilde{R}((s,t),a) \neq \tilde{R}((s,t+1),a) 
\end{align}
The mutable reward structure is dependent on the past, so the properties under the MDP do not hold anymore. However, the policy is invariant~\cite{Ng1999rewardshaping} if two MDPs become graph-isomorphic after a positive linear transformation $f:\mathbb{R} \to \mathbb{R}$ is applied between corresponding weights. Policy invariance also holds when the reward structure is a potential-based shaping function. Reward structure morphing to a policy-invariant MDP preserves the optimal policy, and the value of the current policy. We further define such reward function as a policy-invariant mutable reward structure.

\subsubsection{Mentor Guidance}
Mentor guidance can be interpreted as a policy-invariant mutable reward structure that efficiently trains the learner (mentee).
Parent-child interaction, one of the major mentor guidance in human, is a special form of interaction which establishes a strong bond between the parent and the child and greatly facilitates the cognitive development of a child~\cite{olson1990early}.
%Parent–Child Interaction, https://oxfordre.com/communication/view/10.1093/acrefore/9780190228613.001.0001/acrefore-9780190228613-e-278
The parent agent observes the child's state and has interactions with the child to optimize the child's ability on various tasks. It shows that continuous care and guidance of mentor can hugely facillitate the mentee's performance. Motivated by it, we investigate a specific case of policy-invariant mutable reward structure, e.g. helper reward, aiming to optimize the learner's performance. We also study the behavior cloning, a mutable reward structure that the policy invariance is not known. The guidance $G_T$ on a task $T$ is defined as a reward structure $G_T = \tilde{R}_{T}(s,a)$ where $s \in S$, $a \in A$ with agents performance metric as $J_{T}$.

%In real world, the guidance may be modeled in a multiagent manner. A mentor can either be modeled as an independent agent with a trainable policy or it can be modeled with a fixed policy and mutual actions. For instance, a helper reward in a navigation task can be viewed as a fixed-policy guidance, in which the mentor observes the mentee's distance from the target object and returns the distance as his action, which is then linearly added to the mentee's reward. For another example, through imitation learning we might feed the demonstration to the mentee's observation. This is a mentor's mutual action that drastically changes the mentee's observation probability. 
%Such kind of guidance through \emph{direct} mutual interaction can be easily modeled.
%to facillitate intrinsic motivation to mimic. 
%A guidance through an indirect audiovisual cue, or weak supervision through sparse reward could be the possible policies. We define a guidance policy as an implementation of the parent agent's policy.
%Guiding with indirect sources is much more difficult to model. The mentor and the learner can interact through the intermediate media, e.g. text or videos. This path of guidance does not specifically target the learner in general, and there is a time gap between mentoring and learning. We hypothesize a virtual mentor in this case, associated with the source material.
\begin{figure*}[h!]
    \centering
    \includegraphics[width=\textwidth]{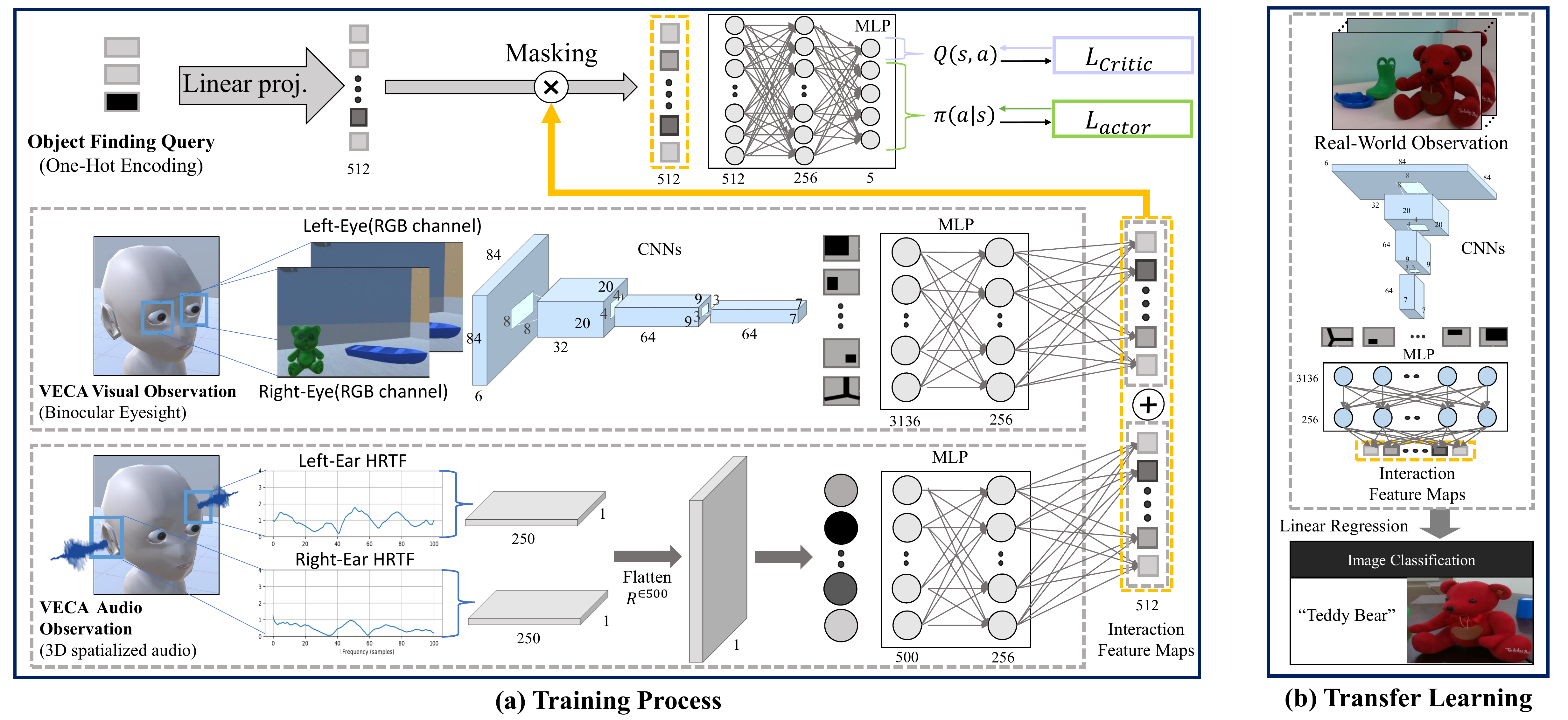}
    \vspace{-20px}
    \caption{Architecture of the baby agent. (a) Its visual observation is encoded as a feature map from the CNN encoder, and the audio observation is encoded with an MLP encoder. These feature maps are concatenated and linearly combined to make an \emph{interaction feature map}. It is masked with a linearly projected object embedding and fed to an MLP, which outputs the Q-function $Q(s,a)$ and movement policy $\pi(a|s)$. (b) We validate helper models in another relevant task via transfer learning}
    \vspace{-10px}
    \label{figure:Agent}
\end{figure*}

\subsection{Critical Period} 
%Researchers in neuroscience and psychology argue that there are critical periods in human learning. 
Critical periods are maturational stages in the lifespan of humans that act as decisive period of learning in a particular learning domain as human grow up~\cite{hensch2004critical}. The critical period for the development of the human visual system is considered to be between three and eight months~\cite{robson2002critical}.  Other critical periods have also been identified for the development of hearing and the vestibular system~\cite{kral2013auditory, brainard1998sensitive}. Besides, critical periods for language acquisition indicate that the first few years of life constitute the time during which language develops readily, and after which 
%(sometime between age 5 and puberty) 
language acquisition is much more difficult.
%and ultimately less successful. 
~\cite{singleton1995agesecondlanguageacquisition} 
%A number of studies show that critical period exists in the development of various cognitive functionalities in a neonatal and infancy period. 
To sum up, critical period stands for the effective initiation time for supervising the cognitive development after birth. We can see the correspondence between the critical period of a toddler and the morphing of a reward structure. In this work, we seek the critical period of an ML agent. We empirically search whether the initial morphing period of best guidance is finite, and can be reached in a reasonable number of training iterations.

% TODO  1/2 Page
\subsection{Optimal Guidance as a Critical Period}
%Guided learning can be defined as a process in which learners achieve their learning goal with guidance from a more experienced mentor or indirect media like text or videos~\cite{billett2000guided}. 
%https://link.springer.com/referenceworkentry/10.1007%2F978-1-4419-1428-6_34
% 
% Source for this?

\subsubsection{Initial Time of Guidance} For human, the parent's guidance initiates from the beginning of the child's life. 
%The training strategies for RL models are usually fixed, or the hyperparameters are annealed to avoid divergence. However, 
For the machine intelligence however, the guidance should persist only for a certain period for a learner to solve the task and learn on their own. We model this characteristic to the RL setting. Given the guidance, the reward morphing starts from an initial time $t_G$ and continues only for a certain amount of time. The guidance resets from this point on. We denote the \textit{initial morphing time} of the guidance as $t_G$. So for a series of learning episodes $\{ E_{i}^G\}$ with guidance $G$ and their duration $\{t_{i}\}$, the time interval of guidance lies in $[t_G, t_G +  \sum_{i} t_{i}]$. This raises the question: What is the optimal initial period of guidance for the learner's performance? 

 %It is evident that the guidance is necessary for the agent to both learn correctly and converge faster. 

 %The infinite guidance may show better performance than any finite-duration guidance. In this case, the optimal guidance duration will be $\infty$. However, finite duration might be sufficient in case the guidance after the duration is redundant or the self-supervising shows better performance after it. 

\subsubsection{Optimal Guidance} We can define an optimal guidance policy on a set of guidance reward structures $\{G_{j}\}$ as the one which results in the best performance with their optimal initial morphing time. Suppose the $J_{T}(\pi_{\theta})$ is the performance of policy parameter $\theta$ on task $T$, and $\Pr(\theta | G, t_{G})$ is a trained policy distribution under the guidance $R = G \in \{G_{j}\}$ and its duration $t_{G}$.  The optimal guidance can be defined as the guidance and its initial morphing period tuple $(G^{*},t_G^{*})$ providing the best performance on a specific task $T$.
\begin{align}
(G^{*}, t_G^{*}) = \arg \max_{G,t_{G}}\bigg(\max_{\theta \sim Pr(\theta | G, t_{G})}\mathbb{E}_{\pi_\theta} \big( J_{T}(\pi_{\theta}) \big) \bigg)
\end{align}
% Guidance policy and practice.

\subsection{Transfer Learning}
Transfer learning aims to help improve learning of some task in the target domain using knowledge learned from the source domain~\cite{weiss2016transfersurvey}.
Transfer learning is used for few-shot learning on new datasets or in previously unseen domains both to reduce training effort and to transfer some of the knowledge and inductive biases acquired during training in the source domain.
%in order to  The target dataset or training algorithm can be different for the source domain and the target domain of the transfer learning, and the target task to transfer is .
One recent work modeled learning-through-play of a toddler by training a Deep Neural Network (DNN) model in a RL framework and transfer this model to a different domain to evaluate it in terms of visual object understanding~\cite{parklearning}. They aim to acquire a general understanding of objects by exploring the environment and actively interacting with the objects, as a toddler does.

\begin{table}[h!]
\begin{tabular}{lll}
\hline
\textbf{Hyperparameter}                              & \textbf{Candidate values}          & \textbf{Optimal value} \\ \hline
$\alpha$ (entropy coefficient) & \{0.003, 0.01, 0.03\}     & 0.01          \\
Learning rate                               & \{0.0001, 0.0003, 0.001\} & 0.0003        \\
$\gamma$ (discount factor)     & \{0.95, 0.99\}            & 0.99         \\ \hline
\end{tabular}
\caption{Tuned hyperparameters used in SAC.}
\vspace{-20px}
\label{table:hyperparam}
\end{table}

\section{Experiments and Results}

\subsection{Implementation Details}
We designed the architecture of the agent as seen in Fig.~\ref{figure:Agent} (a) to learn and store transferable knowledge. We used Soft Actor-Critic (SAC)~\cite{haarnoja2018soft} to train the agent. In particular, we used 8 workers and updated the parameter per 256 steps with batch size 512 using the replay buffer size of 20,000. Other important hyperparameters of SAC are tuned as shown in Table. \ref{table:hyperparam}.
%In particular, we used 0.99 as $\gamma$(discount factor) and updated the parameter per 256 steps with batch size 512 using the replay buffer size of 2e4.

Visual observations of the agent are encoded using a Convolutional Neural Network (CNN) encoder and an Multi-Layer Perceptron (MLP), resulting in \emph{interaction feature maps}. 
These features are masked with a linearly embedded representation of the object and determine the action of the agent. 
Since the agent's movement only depends on masked features, the agent must learn to represent abstract features of its observation corresponding to the target object.
Then the visual feature extractor of the agent is transferred to the EAVE dataset and learned with the Adam~\cite{Adam} gradient descent optimizer. We trained the network on a server with Ubuntu 18.04 LTS, Xeon Gold 5128 Scalable CPU and six RTX3090 GPUs.

\begin{figure}[h!]
    \centering
    \includegraphics[width=0.45\textwidth]{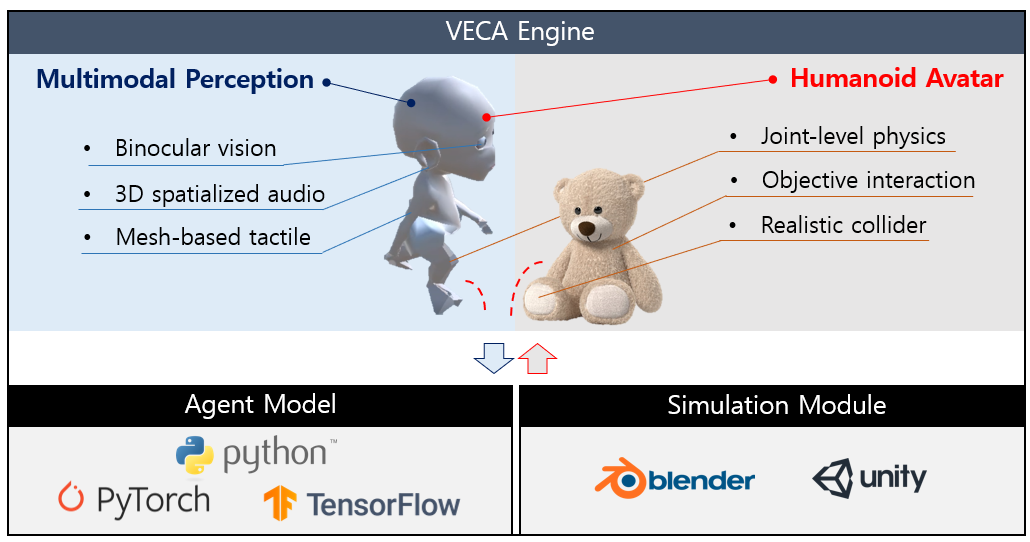}
    \caption{VECA engine characteristics. Unlike other existing RL environments, we trained the toddler AI agents in a setting endowed with various human features and characteristics (e.g. binocular vision).}
     \vspace{-8px}
    \label{figure:VECA}
\end{figure}

\subsection{Interactive Virtual Environment} 
\begin{figure}[h!]
    \centering
    \includegraphics[width=0.5\textwidth]{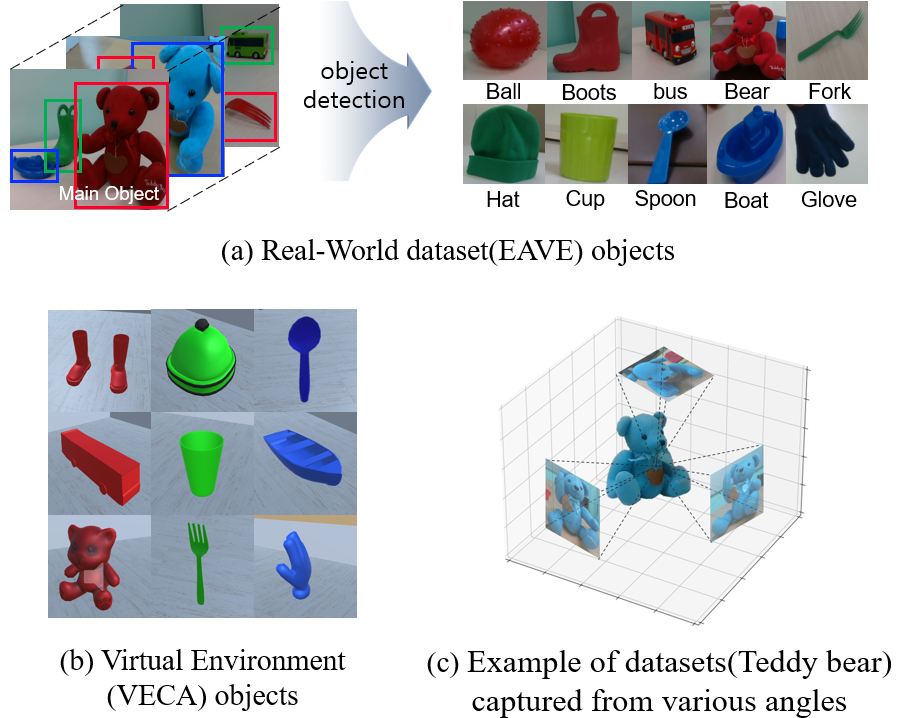}
    \caption{(a) Example of visual images with bounding boxes and the 10 objects used in the EAVE dataset, (b) The same 10 objects in the Virtual Environment (VECA), (c) The EAVE visual dataset includes egocentric images from the toddler's POV at various angles.}
     \vspace{-11px}
    \label{figure:EAVE}
\end{figure}

\begin{figure}[h!]
    \centering
    \includegraphics[width=0.5\textwidth]{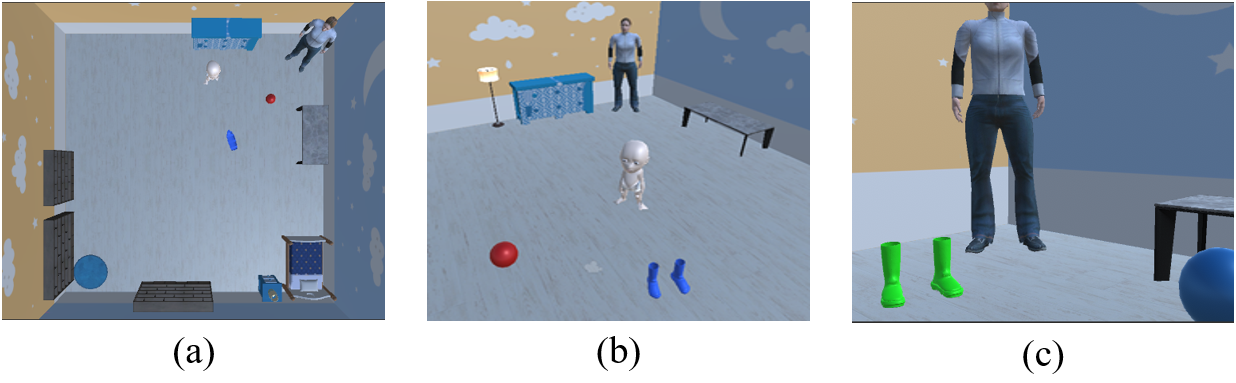}
     \vspace{-8px}
    \caption{Various viewpoints of our 3D environment. (a) Developed 3D-toddler playroom Environment top view, (b) Third-person point of view, (c) First-person point of view of a toddler agent while learning the objects.}
    \vspace{-13px}
    \label{figure:ENV}
\end{figure}

We implemented an interactive virtual environment supporting toddler-like visual observations and active physical interaction with objects, to train and evaluate guidance on different levels and for different durations.  We used VECA toolkit~\cite{VECA}, a virtual environment generation toolkit for human-like agents, to build the toddler-guidance learning environment. In particular, we leveraged a toddler AI agent to be a humanoid avatar including several human characteristics: 1) binocular vision, 2) 3D spatialized audio, 3) mesh-based tactiles, 4) joint-level physics, 5) objective interaction, and a 6) realistic collider as seen in Fig.~\ref{figure:VECA}. The concept of the critical period was founded on human learning. Therefore, we believe that a humanoid agent is able to more precisely learn from observations than most of the existing RL environment agents, which don't reflect the human features for testing the critical period~\cite{kolve2017ai2thor,xia2018gibsonenv,savva2019habitat,chen2020soundspaces}. The agent's goal is to learn the ability to visually locate distant objects, which are the same 10 objects as in the EAVE real-world dataset seen in Fig. \ref{figure:EAVE} (b), while freely exploring and observing nearby objects as seen in Fig~\ref{figure:ENV} (b) and (c). The environment's reward structure is modified to implement the different levels and durations of guidance. We aim to observe the influence of guidance on establishing a profound visual understanding of an object during the RL stage.  

\subsection{EAVE Dataset Collection}
We have collected an Egocentric Audio-Visual Exploration Dataset for Unsupervised Alignment Learning (EAVE dataset) in order to approximate everyday images observed by toddlers. Most existing visual image datasets~\cite{lin2014coco} do not take into account how humans actually perceive the input, meaning their ``first-person'' perspective, thus they cannot fully capture the features of real human-visual input. In particular, the way toddlers perceive the world has two main unique aspects: 1) in the input observed by toddlers objects occupy a relatively larger size compared to an adult's view of the scene (so the image has a much higher resolution); 2) the world is observed by toddlers from a larger variety of different angles than by adults~\cite{bambach2018toddlerinspiredvisualobject}. %When we do the evaluation process with the pre-trained models from the three different levels of social-interaction learning, The input that sufficiently reflects those toddler view's character could enhance our qualitative analysis.
To complement the aforementioned characteristics of the toddler's point of view, we used 10 toys in red, green and blue (for a total of 30 objects) as can be seen in Fig. \ref{figure:EAVE} (a)
and collected 30,000 images using a depth camera. Because toddlers do not typically see only a single object in real-world environments, each image includes three objects much like the ``real-world observations'' from Fig. \ref{figure:EAVE} (a). We are attempting to collect the image perspectives by assuming the baby's point of view from a variety of angles Fig. \ref{figure:EAVE} (c). Moreover, each image has one main object which has the baby's attention. For each image, we label the bounding box information of the three objects and the main object such that it is possible to crop each object individually as a single-object image. Note that the EAVE dataset is only used for transfer learning experiments in the Study 3.

\begin{figure*}[t!]
    \centering
    \includegraphics[width=\textwidth]{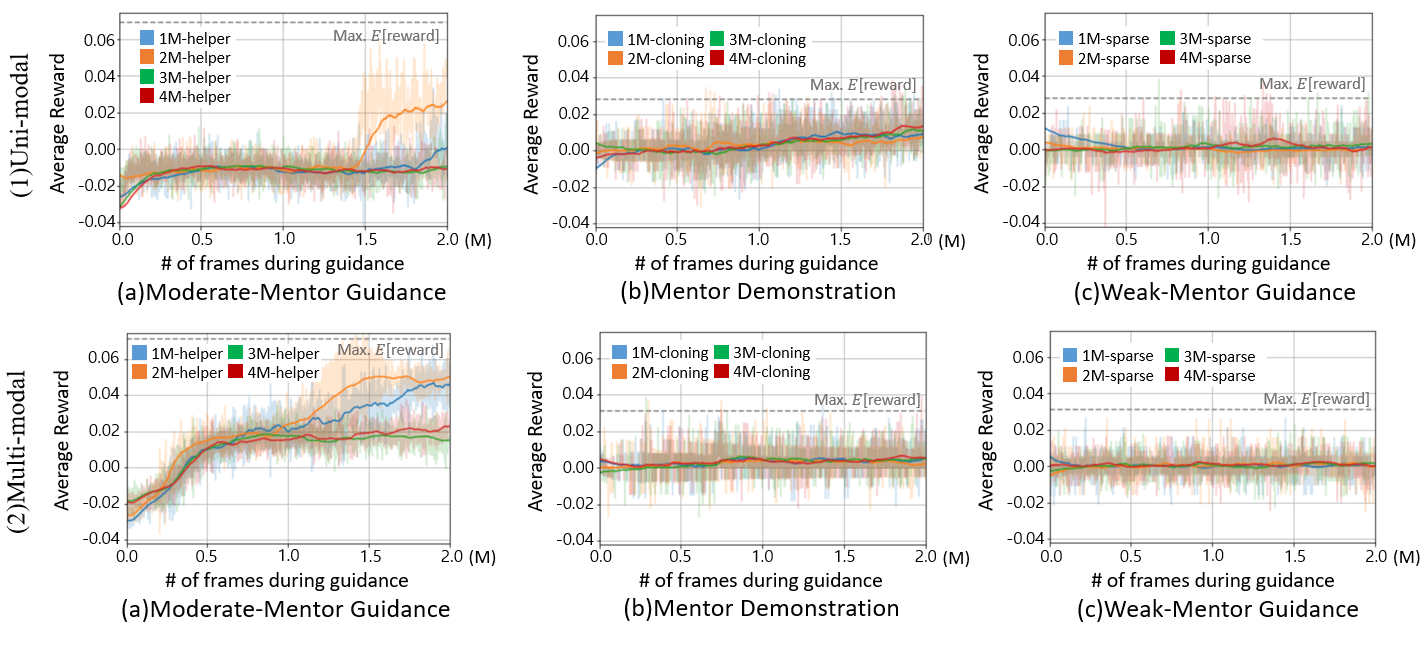}
    \vspace{-25px}
    \caption{Performance of the agents during the guidance. Vertical axis indicates the average reward while the gray dotted line indicates average reward of the ideal policy (0.069/0.028 and 0.077/0.031 for unimodal and multimodal environment with/without helper reward). Horizontal axis indicates the number of frames that the guidance is provided. Agents guided with helper-reward show a remarkable difference in the performance. Agents guided with behavior-cloning improves, but the difference is negligible. Without guidance, agents fail to improve. For helper-reward in multi-modal environment, we used two seeds to ensure that the result is not by the randomness of training. For other experiments, we used single seed.}
    \label{fig:resultG1}
    \vspace{-10px}
\end{figure*}

\subsection{Study 1: Examining the impact of the level of interaction in unimodal learning}

The purpose of study1 is to compare the performance of the agent in a visual (unimodal) environment trained with various start times (for a fixed duration of guidance) and mutual interaction in the form of guidance, and find if there exists a critical period for each mutual interaction. To observe the effect of mutual interaction, we designed the procedure of the experiment as shown in the left side of Fig. \ref{figure:Agent}. We first trained the agent for 1/2/3/4 million frames only with sparse rewards. For each pre-trained agent, we then continued training the agent with guidance (helper-reward, behavior cloning) or without guidance for 2 million frames as seen in Fig~\ref{figure:overallapproach}.

\subsubsection{Task Design} We designed an object-finding task to train the agent in the virtual environment. We used the VECA~\cite{VECA} toolkit to implement and built up an interactive 3D environment that includes ten objects (of the same categories as in the EAVE dataset) and a baby agent. For each episode, two objects are colored randomly (in red, green, blue) and randomly placed with the agent in the square room with length of 18 units, with the distance of at least 4 units for each other as seen in Fig~\ref{figure:ENV} (a). The agent receives the reward +1 for reaching the target object, and receives -1 for reaching the wrong object. We say that the agent reached the object if the agent is looking towards the object and the distance is less than 2.5 units. Note that the distance is necessary because the object, which is on the floor, will disappear from the eyesight if it gets closer since the agent can't move its head vertically. The maximum length of each episode is 256. The agent receives RGB 84x84 binocular vision and receives the information about the target object as a 10-dimensional one-hot vector. The agent can walk around the environment with maximum speed of 0.33 units per frame, which is represented by two continuous action variables $a_f$ (within [0, 1], forward walking speed) and $a_r$ (within [-1, 1], direction of walking). 

\begin{figure*}[t!]
    \centering
    \includegraphics[width=0.95\textwidth]{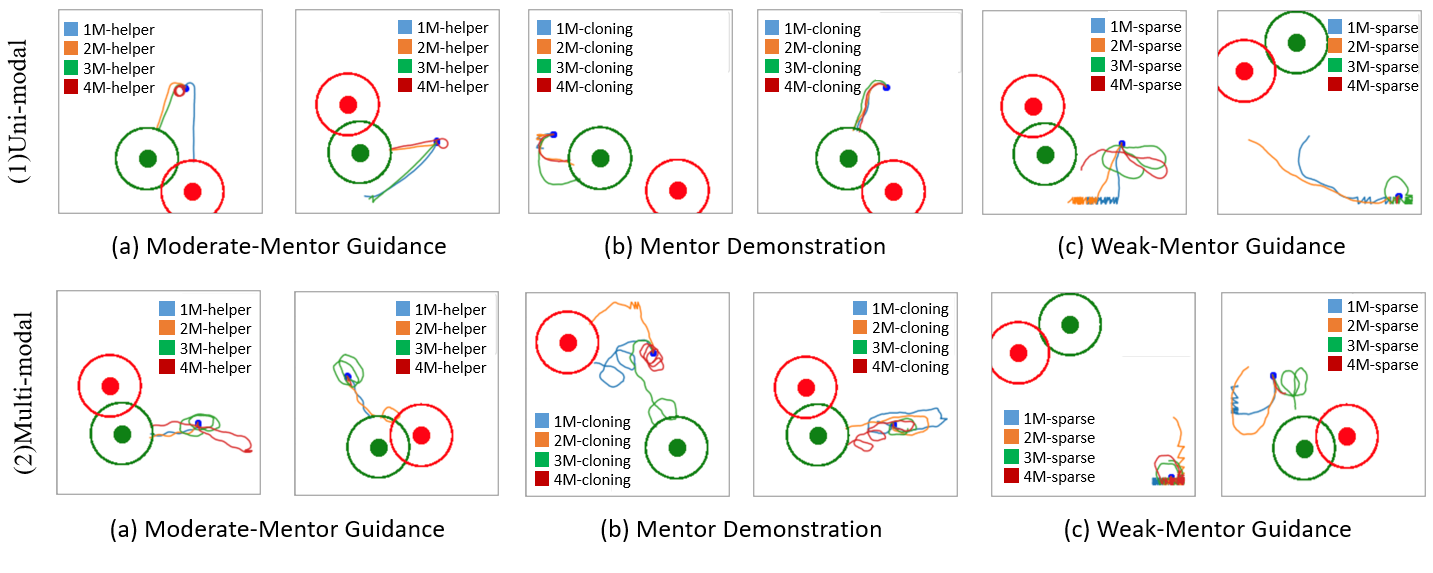}
    \vspace{-15px}
    \caption{Visualization of the trajectory for each agent's policy in unimodal and multimodal environments. Blue dots represent starting points of the agents. Green/red dots and circles represent the target/wrong object and the boundary for reaching it respectively. Note that the agent has to get inside the boundary and also face towards the object to clear the task.}
    \vspace{-8px}
    \label{fig:resultGM2}
\end{figure*}

Please note that this task is not easy to solve with SAC without guidance. First, random (maximum-entropy) actions in this environment are likely to lead the agent to head into the wall, which fills most of the replay buffer with unhelpful data. 
Moreover, the agent has to learn color- and  perspective-independent features of the object since the object is placed and colored randomly. Also, the presence of a wrong object requires the agent not only to navigate to the target object, but also to avoid the wrong object.

This reward function (3) encourages the agent to look towards the object, visually recognize it, and move towards the object. 

\subsubsection{Design of guidance} For moderate mentor guidance, we gave the agent helper reward formulated as follows:
\begin{equation}
r^{'} = \begin{cases}
    -0.03 a_f             &\text{If target is out of eyesight}\\
    0.05 a_r + 0.03 a_f  &\text{If target is left of eyesight} \\
    -0.05 a_r + 0.03 a_f  &\text{If target is right of eyesight} \\
\end{cases}
\end{equation}

For mentor demonstration, we set the mentor's policy according to its visual information. In particular, if the target object is out of eyesight then the agent turns right to find the object. If the target object is in its eyesight then the agent walks forward and turns slightly left/right if the object is on the right/left side of eyesight.

\subsubsection{Results} For guidance from helper-rewards, we found that there exists a noticeable difference in performance depending on the time from which the guidance is provided as shown in Fig. \ref{fig:resultG1}(1)(a). For convenience, we denote $n$M-helper as the model that is trained with helper reward guidance starting from $n$ million frames. In detail, the 2M-helper had much higher average reward (0.03) than the other agents (0.00/-0.01/-0.01 for 1M/3M/4M). This shows that a critical regime indeed exists for guidance from helper-rewards in the object-finding task. Note that for guidance from helper-rewards, helper reward is included when calculating the reward plotted in  Fig. \ref{fig:resultG1}(1)(a) so the reward is negative on the early stage. Fig. \ref{fig:resultGM2}(1)(a) compares the trajectories of the agents, visually showing that the policy of the 2M-helper performs better than the other agents.

However, we found that there was no remarkable difference between the performance for behavior-cloning. As shown in Fig. \ref{fig:resultG1}(1)(b), the average rewards are similar (around 0.01) for all agents. These agents learned how to navigate to the target object, but could not learn how to avoid the wrong object as shown in Fig. \ref{fig:resultGM2}(1)(b).

Without guidance, the agent was unable to learn to solve the task properly as shown in Fig. \ref{fig:resultG1}(1)(c) and Fig. \ref{fig:resultGM2}(1)(c). Compared to the results with guidance, these experiments show that guidance does aid in the agents' learning process.
%and there exists a critical period for helper-reward guidance.

\subsection{Study 2: Examining the impact of the level of interaction in multimodal learning}

The purpose of study2 is to examine the effect of interaction in a multimodal environment and observe how the modality of the agent affected its performance and the critical period. To do so, we modified the object-finding task in Study 1 to an audio-visual task. Specifically, each object has its unique sound and makes this sound during the whole episode. Since the audio observation of the agent is 3D-spatialized, the agent can recognize the direction of each sound source. Thus, we expect that this task would be easier than Study 1, but still challenging since the agent has to learn the relation between the object and the sound. Other detailed setups about the learning process and guidance are identical to Study 1.

\subsubsection{Results} 

Similar to Study 1, we find that there exists a noticeable difference in performance for guidance from helper-rewards, as shown in Fig. \ref{fig:resultG1}(2)(a). This shows that a critical regime also exists for guidance from helper-rewards in the multimodal object-finding task. However, the critical regime differed from Study 1: the 1M- and 2M-helper had much higher average reward (0.046 and 0.050) than 3M-(0.02), 4M-(0.02) agents. Fig. \ref{fig:resultGM2}(2)(a) compares the trajectories of the agents, visually showing that the policy of the 1M- and 2M-helper performs better than the other agents. We observed that the agent majorly uses the audio to get the direction of the object, so it turns right/left even if the object is slightly left/right, which results in wiggling behavior.
%(Note that this does not affect the performance because forward speed is independent of side speed).

The agents guided with behavior-cloning had only a negligible difference in performance as shown in Fig. \ref{fig:resultG1}(2)(b). These agents showed behavior similar to the agents in Study 1: they learned to navigate to the target object but didn't learn to avoid the wrong object as shown in Fig. \ref{fig:resultGM2}(2)(b). However, the average reward was lower than in the unimodal environment. We think that because the target policy is only based on visual observation, so the audio observation worked as noise during behavior cloning.

Without guidance, the agent couldn't solve the task properly as shown in Fig. \ref{fig:resultG1}(2)(c) and Fig. \ref{fig:resultGM2}(2)(c). These results show that the multi-modality of the agent didn't give rise to a new critical period for guidance methods without critical regime, but affected the time of critical period of the guidance with a critical regime.

\subsection{Study 3: Transfer to a real-world dataset}

In study1, we trained with three levels of guidance learning (Sparse-reward, helper-reward, and behavioral cloning) Fig.~\ref{figure:guidance}. This leads to three distinct sets of CNN and MLP weights from the trained models, and we can evaluate each of these sets of weights in another relevant task via transfer learning as seen in Fig.~\ref{figure:Agent} (b).

We previously observed that the 2M- and 1M-helper models showed a noticeable difference in performance compared to the models with no critical period (3M-, 4M-helper) on both uni- and multimodal AI agents. If such a disparity in performance is valid, we believe that the models pre-trained with helper-reward guidance should show comparable results on real-world data input which better reflects the characteristics of toddlers' observations.

To verify the validity of our claims, we evaluated the transfer performance of image encoders (consisting of the CNN and the MLP) from 1M-, 2M-, 3M-, and 4M-helper models. Specifically, we fixed the parameters of the image encoders after pre-training and connected additional linear layers for image classification.
Next, we used the EAVE dataset images, which was collected in the real world in accordance with the toddler's viewpoint, as input (size 84x84) with a total of 6 channels, consisting of the toddler agent's binocular vision with RGB channels each to the right and left side.
We trained the model for 500\,000 iterations with batch size 32.

\begin{figure}[t]
    \centering
    \includegraphics[width=0.4\textwidth]{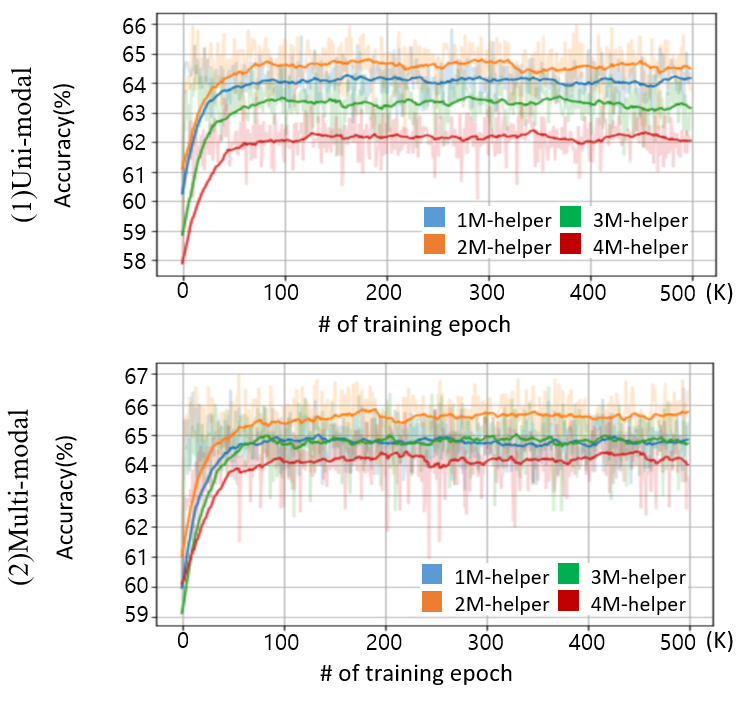}
   % \vspace{-10px}
    \caption{Transfer learning performance with the moderate guidance learning (helper-reward) using the EAVE dataset on unimodal (1) and multimodal (2) agents.}
    \vspace{-10px}
    \label{figure:Transfer}
\end{figure}

As shown in Fig.~\ref{figure:Transfer}, we found that the transferred 2M- ( Uni:64.6/ Multi:65.8\%) and 1M-helper ( Uni:64/Multi:64.9\%) models with a critical period achieved a better performance in terms of accuracy than the 3M- (Uni:63.3/multi:64.9\%) and 4M-helper ( Uni:62.1/multi:64.2\%) models without a critical period on the EAVE dataset.  The multimodal agent's set of CNN and MLP weights has shown a much higher average accuracy than the unimodal agent. Hence, we confirm that multimodal learning rather than unimodal can be helpful in efficiently learning the representation of objects during learning. Furthermore, similar to study1 (see Fig.~\ref{fig:resultG1}~(1),(2)-(a)). we see a distinct difference in moderate guidance learning performance with a helper-reward and can accordingly rank the models in the same order where the models (2M,1M) with a critical period significantly outperform the ones without. Through this, we validate that the learning performance of toddler AI agents indeed depends on the critical period for the setting with moderate guidance learning.

\section{Discussions}

Our paper has made three key contributions as a first step toward discovering the optimal guidance corresponding to critical periods in AI agents. First, to prove that RL-based humanoid AI agents also have critical periods like a toddler has, we found for the first time that 
%much like critical periods in toddler development 
AI agents also show specific critical periods (2M, 1M-helper) within the only moderate guidance learning setting (see Fig.~\ref{fig:resultG1}~(1),(2)-(a)). We assume that 3M/4M frames are past the critical period, so the policy overfits more to the sparse reward, unable to reach the performance of the 2M/1M model.  Behavioral Cloning merely imitates the mentor’s trajectory, unable to generalize beyond the mentor-given policy. However, the helper reward is the optimal guidance among them.

Second, we collected the EAVE dataset, which contains images that imitate a toddler's point of view and its special characteristics and which, along with the developed virtual environment with a humanoid agent, allowed us to verify the qualitative notion of critical periods in a computational and cognitive way. Furthermore, we specifically designed the agent as a humanoid agent, mimicking the toddler's viewpoint's characteristics in VECA, 
so that a humanoid agent can closely learn and more efficiently observe the features of objects in the same way a toddler does. We believe that those human-like learning features enable us to test and validate the critical learning period more reliably.

Lastly, Our findings are especially significant for both Human-AI interaction domain and Human-in-the-loop RL, possibly used to determine the best time for human intervention in the training of AI agents. We developed and designed both uni- and multimodal agents, and confirm that the multimodal agent achieves a considerable improvement over every result of the unimodal agent, inspired by actual human cognitive development. The multimodal agent has a higher average reward and accuracy on object-finding task ( Fig.~\ref{fig:resultG1}~(2)-(a)) and transfer learning on the EAVE dataset (Fig.~\ref{figure:Transfer}~(b)). We validate that multimodal learning at 2M\&1M accompanying the critical period is more effective for learning useful representation compared to the unimodal approach.

In the future, we aim to analyze the critical period based on a more toddler-like RL algorithm deeper in the multi-task\&meta-RL domain or develop an algorithm that automatically finds the critical learning
period of the agent with additional verification studies. In this work, however, we employed a conventional RL algorithm with a light deep learning structure as a first step towards this goal. 

Also, testing the generalizability of the critical period concept requires further experiments. Since the concept is from humans, we speculate that physically grounded humanoid agents will more likely show critical periods here. We plan to study this in various environments including non-humanoid agent in the future.

\section{Conclusion}

Toddlers usually have a distinctive increase in performance during the critical period~\cite{robson2002critical}. At this time, the increase in learning performance depends on the timing of mutual interaction. To investigate a similar effect in learning in AI agents, we studied three levels of mutual interaction learning (forms of guidance) in the virtual environment based on our formulation of toddler guidance learning terms in the RL framework. We validated our observations of the model's critical period on real-world observations (EAVE dataset). Corresponding to the critical period in toddlers, we observed a similar increase in performance using optimal guidance during the training of an AI agent. Specifically, we found that only moderate mutual interaction with a multimodal agent leads to a distinct increase in performance within the critical period and further affects the period during which the AI agent achieves remarkable performance compared to the unimodal agent.

\begin{acks}
This work was partly supported by the IITP (2015-0-00310-SW.Star-Lab/16\%, 2017-0-01772-VTT/16\%, 2018-0-00622-RMI/16\%, 2019-0-01371-BabyMind/20\%) grants and the NRF of Korea (2021R1A2C10-10970/16\%) grant funded by the Korean government, and the CARAI (UD190031RD/16\%) grant funded by the DAPA and ADD.

\end{acks}

%%
%% The next two lines define the bibliography style to be used, and
%% the bibliography file.
\bibliographystyle{ACM-Reference-Format.bst}
\bibliography{acmart.bib}

%%% -*-BibTeX-*-
%%% Do NOT edit. File created by BibTeX with style
%%% ACM-Reference-Format-Journals [18-Jan-2012].

\begin{thebibliography}{56}

%%% ====================================================================
%%% NOTE TO THE USER: you can override these defaults by providing
%%% customized versions of any of these macros before the \bibliography
%%% command.  Each of them MUST provide its own final punctuation,
%%% except for \shownote{}, \showDOI{}, and \showURL{}.  The latter two
%%% do not use final punctuation, in order to avoid confusing it with
%%% the Web address.
%%%
%%% To suppress output of a particular field, define its macro to expand
%%% to an empty string, or better, \unskip, like this:
%%%
%%% \newcommand{\showDOI}[1]{\unskip}   % LaTeX syntax
%%%
%%% \def \showDOI #1{\unskip}           % plain TeX syntax
%%%
%%% ====================================================================

\ifx \showCODEN    \undefined \def \showCODEN     #1{\unskip}     \fi
\ifx \showDOI      \undefined \def \showDOI       #1{#1}\fi
\ifx \showISBNx    \undefined \def \showISBNx     #1{\unskip}     \fi
\ifx \showISBNxiii \undefined \def \showISBNxiii  #1{\unskip}     \fi
\ifx \showISSN     \undefined \def \showISSN      #1{\unskip}     \fi
\ifx \showLCCN     \undefined \def \showLCCN      #1{\unskip}     \fi
\ifx \shownote     \undefined \def \shownote      #1{#1}          \fi
\ifx \showarticletitle \undefined \def \showarticletitle #1{#1}   \fi
\ifx \showURL      \undefined \def \showURL       {\relax}        \fi
% The following commands are used for tagged output and should be
% invisible to TeX
\providecommand\bibfield[2]{#2}
\providecommand\bibinfo[2]{#2}
\providecommand\natexlab[1]{#1}
\providecommand\showeprint[2][]{arXiv:#2}

\bibitem[\protect\citeauthoryear{Achille, Rovere, and Soatto}{Achille
  et~al\mbox{.}}{2018}]%
        {achille2018criticalperiodDNN}
\bibfield{author}{\bibinfo{person}{Alessandro Achille}, \bibinfo{person}{Matteo
  Rovere}, {and} \bibinfo{person}{Stefano Soatto}.}
  \bibinfo{year}{2018}\natexlab{}.
\newblock \showarticletitle{Critical learning periods in deep networks}. In
  \bibinfo{booktitle}{\emph{International Conference on Learning
  Representations}}.
\newblock


\bibitem[\protect\citeauthoryear{Arulkumaran, Deisenroth, Brundage, and
  Bharath}{Arulkumaran et~al\mbox{.}}{2017}]%
        {arulkumaran2017briefRLSurvey}
\bibfield{author}{\bibinfo{person}{Kai Arulkumaran},
  \bibinfo{person}{Marc~Peter Deisenroth}, \bibinfo{person}{Miles Brundage},
  {and} \bibinfo{person}{Anil~Anthony Bharath}.}
  \bibinfo{year}{2017}\natexlab{}.
\newblock \showarticletitle{A brief survey of deep reinforcement learning}.
\newblock \bibinfo{journal}{\emph{arXiv preprint arXiv:1708.05866}}
  (\bibinfo{year}{2017}).
\newblock


\bibitem[\protect\citeauthoryear{Bambach, Crandall, Smith, and Yu}{Bambach
  et~al\mbox{.}}{2018}]%
        {bambach2018toddlerinspiredvisualobject}
\bibfield{author}{\bibinfo{person}{Sven Bambach}, \bibinfo{person}{David
  Crandall}, \bibinfo{person}{Linda Smith}, {and} \bibinfo{person}{Chen Yu}.}
  \bibinfo{year}{2018}\natexlab{}.
\newblock \showarticletitle{Toddler-inspired visual object learning}. In
  \bibinfo{booktitle}{\emph{Advances in neural information processing
  systems}}. \bibinfo{pages}{1201--1210}.
\newblock


\bibitem[\protect\citeauthoryear{Bengio, Louradour, Collobert, and
  Weston}{Bengio et~al\mbox{.}}{2009}]%
        {Bengio2009CurriculumLearning}
\bibfield{author}{\bibinfo{person}{Yoshua Bengio}, \bibinfo{person}{J.
  Louradour}, \bibinfo{person}{Ronan Collobert}, {and} \bibinfo{person}{J.
  Weston}.} \bibinfo{year}{2009}\natexlab{}.
\newblock \showarticletitle{Curriculum learning}. In
  \bibinfo{booktitle}{\emph{ICML '09}}.
\newblock


\bibitem[\protect\citeauthoryear{Billett}{Billett}{2000}]%
        {billett2000guided}
\bibfield{author}{\bibinfo{person}{Stephen Billett}.}
  \bibinfo{year}{2000}\natexlab{}.
\newblock \showarticletitle{Guided learning at work}.
\newblock \bibinfo{journal}{\emph{Journal of Workplace learning}}
  (\bibinfo{year}{2000}).
\newblock


\bibitem[\protect\citeauthoryear{Brainard and Knudsen}{Brainard and
  Knudsen}{1998}]%
        {brainard1998sensitive}
\bibfield{author}{\bibinfo{person}{Michael~S Brainard} {and}
  \bibinfo{person}{Eric~I Knudsen}.} \bibinfo{year}{1998}\natexlab{}.
\newblock \showarticletitle{Sensitive periods for visual calibration of the
  auditory space map in the barn owl optic tectum}.
\newblock \bibinfo{journal}{\emph{Journal of Neuroscience}}
  \bibinfo{volume}{18}, \bibinfo{number}{10} (\bibinfo{year}{1998}),
  \bibinfo{pages}{3929--3942}.
\newblock


\bibitem[\protect\citeauthoryear{Braun, Neil, and Liu}{Braun
  et~al\mbox{.}}{2017}]%
        {Braun2017ACL}
\bibfield{author}{\bibinfo{person}{Stefan Braun}, \bibinfo{person}{Daniel
  Neil}, {and} \bibinfo{person}{Shih-Chii Liu}.}
  \bibinfo{year}{2017}\natexlab{}.
\newblock \showarticletitle{A curriculum learning method for improved noise
  robustness in automatic speech recognition}.
\newblock \bibinfo{journal}{\emph{2017 25th European Signal Processing
  Conference (EUSIPCO)}} (\bibinfo{year}{2017}), \bibinfo{pages}{548--552}.
\newblock


\bibitem[\protect\citeauthoryear{Chen, Jain, Schissler, Gari, Al-Halah, Ithapu,
  Robinson, and Grauman}{Chen et~al\mbox{.}}{2020}]%
        {chen2020soundspaces}
\bibfield{author}{\bibinfo{person}{Changan Chen}, \bibinfo{person}{Unnat Jain},
  \bibinfo{person}{Carl Schissler}, \bibinfo{person}{Sebastia Vicenc~Amengual
  Gari}, \bibinfo{person}{Ziad Al-Halah}, \bibinfo{person}{Vamsi~Krishna
  Ithapu}, \bibinfo{person}{Philip Robinson}, {and} \bibinfo{person}{Kristen
  Grauman}.} \bibinfo{year}{2020}\natexlab{}.
\newblock \showarticletitle{Soundspaces: Audio-visual navigation in 3d
  environments}. In \bibinfo{booktitle}{\emph{Proceedings of the European
  Conference on Computer Vision (ECCV)}}. Springer.
\newblock


\bibitem[\protect\citeauthoryear{Chen and Gupta}{Chen and Gupta}{2015}]%
        {chen2015weblyCL}
\bibfield{author}{\bibinfo{person}{Xinlei Chen} {and} \bibinfo{person}{Abhinav
  Gupta}.} \bibinfo{year}{2015}\natexlab{}.
\newblock \showarticletitle{Webly supervised learning of convolutional
  networks}. In \bibinfo{booktitle}{\emph{Proceedings of the IEEE International
  Conference on Computer Vision}}. \bibinfo{pages}{1431--1439}.
\newblock


\bibitem[\protect\citeauthoryear{Chiswick and Miller}{Chiswick and
  Miller}{2008}]%
        {chiswick2008testcriticalperiod}
\bibfield{author}{\bibinfo{person}{Barry~R Chiswick} {and}
  \bibinfo{person}{Paul~W Miller}.} \bibinfo{year}{2008}\natexlab{}.
\newblock \showarticletitle{A test of the critical period hypothesis for
  language learning}.
\newblock \bibinfo{journal}{\emph{Journal of multilingual and multicultural
  development}} \bibinfo{volume}{29}, \bibinfo{number}{1}
  (\bibinfo{year}{2008}), \bibinfo{pages}{16--29}.
\newblock


\bibitem[\protect\citeauthoryear{Darley and Fazio}{Darley and Fazio}{1980}]%
        {darley1980expectancysocialinteraction}
\bibfield{author}{\bibinfo{person}{John~M Darley} {and}
  \bibinfo{person}{Russell~H Fazio}.} \bibinfo{year}{1980}\natexlab{}.
\newblock \showarticletitle{Expectancy confirmation processes arising in the
  social interaction sequence.}
\newblock \bibinfo{journal}{\emph{American Psychologist}} \bibinfo{volume}{35},
  \bibinfo{number}{10} (\bibinfo{year}{1980}), \bibinfo{pages}{867}.
\newblock


\bibitem[\protect\citeauthoryear{Doshi-Velez}{Doshi-Velez}{2009}]%
        {doshi2009infinitePOMDP}
\bibfield{author}{\bibinfo{person}{Finale Doshi-Velez}.}
  \bibinfo{year}{2009}\natexlab{}.
\newblock \showarticletitle{The infinite partially observable Markov decision
  process}.
\newblock \bibinfo{journal}{\emph{Advances in neural information processing
  systems}}  \bibinfo{volume}{22} (\bibinfo{year}{2009}),
  \bibinfo{pages}{477--485}.
\newblock


\bibitem[\protect\citeauthoryear{Elman}{Elman}{1993}]%
        {Elman1993Learningsmall}
\bibfield{author}{\bibinfo{person}{J. Elman}.} \bibinfo{year}{1993}\natexlab{}.
\newblock \showarticletitle{Learning and development in neural networks: the
  importance of starting small}.
\newblock \bibinfo{journal}{\emph{Cognition}}  \bibinfo{volume}{48}
  (\bibinfo{year}{1993}), \bibinfo{pages}{71--99}.
\newblock


\bibitem[\protect\citeauthoryear{Gibson}{Gibson}{1988}]%
        {gibson1988toddlerlearning}
\bibfield{author}{\bibinfo{person}{Eleanor~J Gibson}.}
  \bibinfo{year}{1988}\natexlab{}.
\newblock \showarticletitle{Exploratory behavior in the development of
  perceiving, acting, and the acquiring of knowledge}.
\newblock \bibinfo{journal}{\emph{Annual review of psychology}}
  \bibinfo{volume}{39}, \bibinfo{number}{1} (\bibinfo{year}{1988}),
  \bibinfo{pages}{1--42}.
\newblock


\bibitem[\protect\citeauthoryear{Haarnoja, Zhou, Abbeel, and Levine}{Haarnoja
  et~al\mbox{.}}{2018}]%
        {haarnoja2018soft}
\bibfield{author}{\bibinfo{person}{Tuomas Haarnoja}, \bibinfo{person}{Aurick
  Zhou}, \bibinfo{person}{Pieter Abbeel}, {and} \bibinfo{person}{Sergey
  Levine}.} \bibinfo{year}{2018}\natexlab{}.
\newblock \showarticletitle{Soft actor-critic: Off-policy maximum entropy deep
  reinforcement learning with a stochastic actor}. In
  \bibinfo{booktitle}{\emph{International Conference on Machine Learning}}.
  PMLR, \bibinfo{pages}{1861--1870}.
\newblock


\bibitem[\protect\citeauthoryear{Hensch}{Hensch}{2004}]%
        {hensch2004critical}
\bibfield{author}{\bibinfo{person}{Takao~K Hensch}.}
  \bibinfo{year}{2004}\natexlab{}.
\newblock \showarticletitle{Critical period regulation}.
\newblock \bibinfo{journal}{\emph{Annu. Rev. Neurosci.}}  \bibinfo{volume}{27}
  (\bibinfo{year}{2004}), \bibinfo{pages}{549--579}.
\newblock


\bibitem[\protect\citeauthoryear{Homans}{Homans}{1974}]%
        {homans1974social}
\bibfield{author}{\bibinfo{person}{George~C Homans}.}
  \bibinfo{year}{1974}\natexlab{}.
\newblock \showarticletitle{Social behavior: Its elementary forms}.
\newblock  (\bibinfo{year}{1974}).
\newblock


\bibitem[\protect\citeauthoryear{Jesson, Guizard, Ghalehjegh, Goblot, Soudan,
  and Chapados}{Jesson et~al\mbox{.}}{2017}]%
        {Jesson2017CASEDCA}
\bibfield{author}{\bibinfo{person}{Andrew Jesson}, \bibinfo{person}{N.
  Guizard}, \bibinfo{person}{Sina~Hamidi Ghalehjegh}, \bibinfo{person}{D.
  Goblot}, \bibinfo{person}{F. Soudan}, {and} \bibinfo{person}{Nicolas
  Chapados}.} \bibinfo{year}{2017}\natexlab{}.
\newblock \showarticletitle{CASED: Curriculum Adaptive Sampling for Extreme
  Data Imbalance}.
\newblock \bibinfo{journal}{\emph{ArXiv}}  \bibinfo{volume}{abs/1807.10819}
  (\bibinfo{year}{2017}).
\newblock


\bibitem[\protect\citeauthoryear{Kingma and Ba}{Kingma and Ba}{2014}]%
        {Adam}
\bibfield{author}{\bibinfo{person}{Diederik~P Kingma} {and}
  \bibinfo{person}{Jimmy Ba}.} \bibinfo{year}{2014}\natexlab{}.
\newblock \showarticletitle{Adam: A method for stochastic optimization}.
\newblock \bibinfo{journal}{\emph{arXiv preprint arXiv:1412.6980}}
  (\bibinfo{year}{2014}).
\newblock


\bibitem[\protect\citeauthoryear{Kolve, Mottaghi, Han, VanderBilt, Weihs,
  Herrasti, Gordon, Zhu, Gupta, and Farhadi}{Kolve et~al\mbox{.}}{2017}]%
        {kolve2017ai2thor}
\bibfield{author}{\bibinfo{person}{Eric Kolve}, \bibinfo{person}{Roozbeh
  Mottaghi}, \bibinfo{person}{Winson Han}, \bibinfo{person}{Eli VanderBilt},
  \bibinfo{person}{Luca Weihs}, \bibinfo{person}{Alvaro Herrasti},
  \bibinfo{person}{Daniel Gordon}, \bibinfo{person}{Yuke Zhu},
  \bibinfo{person}{Abhinav Gupta}, {and} \bibinfo{person}{Ali Farhadi}.}
  \bibinfo{year}{2017}\natexlab{}.
\newblock \showarticletitle{Ai2-thor: An interactive 3d environment for visual
  ai}.
\newblock \bibinfo{journal}{\emph{arXiv preprint arXiv:1712.05474}}
  (\bibinfo{year}{2017}).
\newblock


\bibitem[\protect\citeauthoryear{Kral}{Kral}{2013}]%
        {kral2013auditory}
\bibfield{author}{\bibinfo{person}{A Kral}.} \bibinfo{year}{2013}\natexlab{}.
\newblock \showarticletitle{Auditory critical periods: a review from system’s
  perspective}.
\newblock \bibinfo{journal}{\emph{Neuroscience}}  \bibinfo{volume}{247}
  (\bibinfo{year}{2013}), \bibinfo{pages}{117--133}.
\newblock


\bibitem[\protect\citeauthoryear{Krashen}{Krashen}{1973}]%
        {krashen1973lateralizationcritical}
\bibfield{author}{\bibinfo{person}{Stephen~D Krashen}.}
  \bibinfo{year}{1973}\natexlab{}.
\newblock \showarticletitle{Lateralization, language learning, and the critical
  period: Some new evidence}.
\newblock \bibinfo{journal}{\emph{Language learning}} \bibinfo{volume}{23},
  \bibinfo{number}{1} (\bibinfo{year}{1973}), \bibinfo{pages}{63--74}.
\newblock


\bibitem[\protect\citeauthoryear{Lee, Eysenbach, Salakhutdinov, Gu, and
  Finn}{Lee et~al\mbox{.}}{2020}]%
        {lee2020weaklysupervised}
\bibfield{author}{\bibinfo{person}{Lisa Lee}, \bibinfo{person}{Ben Eysenbach},
  \bibinfo{person}{Russ~R Salakhutdinov}, \bibinfo{person}{Shixiang~Shane Gu},
  {and} \bibinfo{person}{Chelsea Finn}.} \bibinfo{year}{2020}\natexlab{}.
\newblock \showarticletitle{Weakly-Supervised Reinforcement Learning for
  Controllable Behavior}.
\newblock \bibinfo{journal}{\emph{Advances in Neural Information Processing
  Systems}}  \bibinfo{volume}{33} (\bibinfo{year}{2020}).
\newblock


\bibitem[\protect\citeauthoryear{Lin, Wang, Meng, Zuo, and Zhang}{Lin
  et~al\mbox{.}}{2018}]%
        {Lin2018ActiveSL}
\bibfield{author}{\bibinfo{person}{L. Lin}, \bibinfo{person}{Keze Wang},
  \bibinfo{person}{Deyu Meng}, \bibinfo{person}{W. Zuo}, {and}
  \bibinfo{person}{Lei Zhang}.} \bibinfo{year}{2018}\natexlab{}.
\newblock \showarticletitle{Active Self-Paced Learning for Cost-Effective and
  Progressive Face Identification}.
\newblock \bibinfo{journal}{\emph{IEEE Transactions on Pattern Analysis and
  Machine Intelligence}}  \bibinfo{volume}{40} (\bibinfo{year}{2018}),
  \bibinfo{pages}{7--19}.
\newblock


\bibitem[\protect\citeauthoryear{Lin, Maire, Belongie, Hays, Perona, Ramanan,
  Doll{\'a}r, and Zitnick}{Lin et~al\mbox{.}}{2014}]%
        {lin2014coco}
\bibfield{author}{\bibinfo{person}{Tsung-Yi Lin}, \bibinfo{person}{Michael
  Maire}, \bibinfo{person}{Serge Belongie}, \bibinfo{person}{James Hays},
  \bibinfo{person}{Pietro Perona}, \bibinfo{person}{Deva Ramanan},
  \bibinfo{person}{Piotr Doll{\'a}r}, {and} \bibinfo{person}{C~Lawrence
  Zitnick}.} \bibinfo{year}{2014}\natexlab{}.
\newblock \showarticletitle{Microsoft coco: Common objects in context}. In
  \bibinfo{booktitle}{\emph{European conference on computer vision}}. Springer,
  \bibinfo{pages}{740--755}.
\newblock


\bibitem[\protect\citeauthoryear{Lu}{Lu}{2009}]%
        {lu2009semisupervisedlearning}
\bibfield{author}{\bibinfo{person}{Tyler~Tian Lu}.}
  \bibinfo{year}{2009}\natexlab{}.
\newblock \emph{\bibinfo{title}{Fundamental limitations of semi-supervised
  learning}}.
\newblock \bibinfo{thesistype}{Master's\ thesis}. \bibinfo{school}{University
  of Waterloo}.
\newblock


\bibitem[\protect\citeauthoryear{Ng, Harada, and Russell}{Ng
  et~al\mbox{.}}{1999}]%
        {Ng1999rewardshaping}
\bibfield{author}{\bibinfo{person}{A. Ng}, \bibinfo{person}{D. Harada}, {and}
  \bibinfo{person}{Stuart~J. Russell}.} \bibinfo{year}{1999}\natexlab{}.
\newblock \showarticletitle{Policy Invariance Under Reward Transformations:
  Theory and Application to Reward Shaping}. In
  \bibinfo{booktitle}{\emph{ICML}}.
\newblock


\bibitem[\protect\citeauthoryear{Olson, Bates, and Bayles}{Olson
  et~al\mbox{.}}{1990}]%
        {olson1990early}
\bibfield{author}{\bibinfo{person}{Sheryl~L Olson}, \bibinfo{person}{John~E
  Bates}, {and} \bibinfo{person}{Kathryn Bayles}.}
  \bibinfo{year}{1990}\natexlab{}.
\newblock \showarticletitle{Early antecedents of childhood impulsivity: The
  role of parent-child interaction, cognitive competence, and temperament}.
\newblock \bibinfo{journal}{\emph{Journal of abnormal child psychology}}
  \bibinfo{volume}{18}, \bibinfo{number}{3} (\bibinfo{year}{1990}),
  \bibinfo{pages}{317--334}.
\newblock


\bibitem[\protect\citeauthoryear{Park, Heo, and Lee}{Park
  et~al\mbox{.}}{2020}]%
        {VECA}
\bibfield{author}{\bibinfo{person}{Kwanyoung Park}, \bibinfo{person}{Jeong
  Heo}, {and} \bibinfo{person}{Youngki Lee}.} \bibinfo{year}{2020}\natexlab{}.
\newblock \bibinfo{title}{VECA: A VR Toolkit for Training and Testing Cognitive
  Agents}.
\newblock \bibinfo{howpublished}{\url{https://github .com/GGOSinon/VECA}}.
\newblock


\bibitem[\protect\citeauthoryear{Park, Park, Oh, Zhang, and Lee}{Park
  et~al\mbox{.}}{2021}]%
        {parklearning}
\bibfield{author}{\bibinfo{person}{Kwanyoung Park}, \bibinfo{person}{Junseok
  Park}, \bibinfo{person}{Hyunseok Oh}, \bibinfo{person}{Byoung-Tak Zhang},
  {and} \bibinfo{person}{Youngki Lee}.} \bibinfo{year}{2021}\natexlab{}.
\newblock \bibinfo{title}{Learning task-agnostic representation via
  toddler-inspired learning}.
\newblock
\newblock
\showeprint[arxiv]{2101.11221}~[cs.AI]


\bibitem[\protect\citeauthoryear{Pentina, Sharmanska, and Lampert}{Pentina
  et~al\mbox{.}}{2015}]%
        {pentina2015curriculumCL}
\bibfield{author}{\bibinfo{person}{Anastasia Pentina},
  \bibinfo{person}{Viktoriia Sharmanska}, {and} \bibinfo{person}{Christoph~H
  Lampert}.} \bibinfo{year}{2015}\natexlab{}.
\newblock \showarticletitle{Curriculum learning of multiple tasks}. In
  \bibinfo{booktitle}{\emph{Proceedings of the IEEE Conference on Computer
  Vision and Pattern Recognition}}. \bibinfo{pages}{5492--5500}.
\newblock


\bibitem[\protect\citeauthoryear{Pi, Li, Zhang, Meng, Wu, Xiao, and Zhuang}{Pi
  et~al\mbox{.}}{2016}]%
        {Pi2016SelfPacedBL}
\bibfield{author}{\bibinfo{person}{Te Pi}, \bibinfo{person}{Xi Li},
  \bibinfo{person}{Zhongfei Zhang}, \bibinfo{person}{Deyu Meng},
  \bibinfo{person}{Fei Wu}, \bibinfo{person}{Jun Xiao}, {and}
  \bibinfo{person}{Yueting Zhuang}.} \bibinfo{year}{2016}\natexlab{}.
\newblock \showarticletitle{Self-Paced Boost Learning for Classification}. In
  \bibinfo{booktitle}{\emph{IJCAI}}.
\newblock


\bibitem[\protect\citeauthoryear{Piaget and Cook}{Piaget and Cook}{1952}]%
        {piaget1952origins}
\bibfield{author}{\bibinfo{person}{Jean Piaget} {and} \bibinfo{person}{Margaret
  Cook}.} \bibinfo{year}{1952}\natexlab{}.
\newblock \bibinfo{booktitle}{\emph{The origins of intelligence in children}}.
  Vol.~\bibinfo{volume}{8}.
\newblock \bibinfo{publisher}{International Universities Press New York}.
\newblock


\bibitem[\protect\citeauthoryear{Robson}{Robson}{2002}]%
        {robson2002critical}
\bibfield{author}{\bibinfo{person}{Ann~L Robson}.}
  \bibinfo{year}{2002}\natexlab{}.
\newblock \showarticletitle{Critical/sensitive periods}.
\newblock \bibinfo{journal}{\emph{Child Develop-ment}} (\bibinfo{year}{2002}),
  \bibinfo{pages}{101--103}.
\newblock


\bibitem[\protect\citeauthoryear{Savva, Kadian, Maksymets, Zhao, Wijmans, Jain,
  Straub, Liu, Koltun, Malik, et~al\mbox{.}}{Savva et~al\mbox{.}}{2019}]%
        {savva2019habitat}
\bibfield{author}{\bibinfo{person}{Manolis Savva}, \bibinfo{person}{Abhishek
  Kadian}, \bibinfo{person}{Oleksandr Maksymets}, \bibinfo{person}{Yili Zhao},
  \bibinfo{person}{Erik Wijmans}, \bibinfo{person}{Bhavana Jain},
  \bibinfo{person}{Julian Straub}, \bibinfo{person}{Jia Liu},
  \bibinfo{person}{Vladlen Koltun}, \bibinfo{person}{Jitendra Malik},
  {et~al\mbox{.}}} \bibinfo{year}{2019}\natexlab{}.
\newblock \showarticletitle{Habitat: A platform for embodied ai research}. In
  \bibinfo{booktitle}{\emph{Proceedings of the IEEE/CVF International
  Conference on Computer Vision}}. \bibinfo{pages}{9339--9347}.
\newblock


\bibitem[\protect\citeauthoryear{Schank}{Schank}{1972}]%
        {schank1972nlp}
\bibfield{author}{\bibinfo{person}{Roger~C Schank}.}
  \bibinfo{year}{1972}\natexlab{}.
\newblock \showarticletitle{Conceptual dependency: A theory of natural language
  understanding}.
\newblock \bibinfo{journal}{\emph{Cognitive psychology}} \bibinfo{volume}{3},
  \bibinfo{number}{4} (\bibinfo{year}{1972}), \bibinfo{pages}{552--631}.
\newblock


\bibitem[\protect\citeauthoryear{Shi, Larson, and Jonker}{Shi
  et~al\mbox{.}}{2015}]%
        {shi2015recurrentCL}
\bibfield{author}{\bibinfo{person}{Yangyang Shi}, \bibinfo{person}{Martha
  Larson}, {and} \bibinfo{person}{Catholijn~M Jonker}.}
  \bibinfo{year}{2015}\natexlab{}.
\newblock \showarticletitle{Recurrent neural network language model adaptation
  with curriculum learning}.
\newblock \bibinfo{journal}{\emph{Computer Speech \& Language}}
  \bibinfo{volume}{33}, \bibinfo{number}{1} (\bibinfo{year}{2015}),
  \bibinfo{pages}{136--154}.
\newblock


\bibitem[\protect\citeauthoryear{Shoham, Powers, and Grenager}{Shoham
  et~al\mbox{.}}{2003}]%
        {shoham2003MultiagentRLCriticalSurvey}
\bibfield{author}{\bibinfo{person}{Yoav Shoham}, \bibinfo{person}{Rob Powers},
  {and} \bibinfo{person}{Trond Grenager}.} \bibinfo{year}{2003}\natexlab{}.
\newblock \showarticletitle{Multi-agent reinforcement learning: a critical
  survey}.
\newblock  (\bibinfo{year}{2003}).
\newblock


\bibitem[\protect\citeauthoryear{Shrivastava, Gupta, and Girshick}{Shrivastava
  et~al\mbox{.}}{2016}]%
        {Shrivastava2016TrainingRO}
\bibfield{author}{\bibinfo{person}{Abhinav Shrivastava}, \bibinfo{person}{A.
  Gupta}, {and} \bibinfo{person}{Ross~B. Girshick}.}
  \bibinfo{year}{2016}\natexlab{}.
\newblock \showarticletitle{Training Region-Based Object Detectors with Online
  Hard Example Mining}.
\newblock \bibinfo{journal}{\emph{2016 IEEE Conference on Computer Vision and
  Pattern Recognition (CVPR)}} (\bibinfo{year}{2016}),
  \bibinfo{pages}{761--769}.
\newblock


\bibitem[\protect\citeauthoryear{Singleton and Lengyel}{Singleton and
  Lengyel}{1995}]%
        {singleton1995agesecondlanguageacquisition}
\bibfield{author}{\bibinfo{person}{David~Michael Singleton} {and}
  \bibinfo{person}{Zsolt Lengyel}.} \bibinfo{year}{1995}\natexlab{}.
\newblock \bibinfo{booktitle}{\emph{The age factor in second language
  acquisition: A critical look at the critical period hypothesis}}.
\newblock \bibinfo{publisher}{Multilingual Matters}.
\newblock


\bibitem[\protect\citeauthoryear{Snow and Hoefnagel-H{\"o}hle}{Snow and
  Hoefnagel-H{\"o}hle}{1978}]%
        {snow1978criticalperiodlang}
\bibfield{author}{\bibinfo{person}{Catherine~E Snow} {and}
  \bibinfo{person}{Marian Hoefnagel-H{\"o}hle}.}
  \bibinfo{year}{1978}\natexlab{}.
\newblock \showarticletitle{The critical period for language acquisition:
  Evidence from second language learning}.
\newblock \bibinfo{journal}{\emph{Child development}} (\bibinfo{year}{1978}),
  \bibinfo{pages}{1114--1128}.
\newblock


\bibitem[\protect\citeauthoryear{Soviany, Ardei, Ionescu, and
  Leordeanu}{Soviany et~al\mbox{.}}{2020}]%
        {Soviany2020ImageDC}
\bibfield{author}{\bibinfo{person}{Petru Soviany}, \bibinfo{person}{Claudiu
  Ardei}, \bibinfo{person}{Radu~Tudor Ionescu}, {and} \bibinfo{person}{M.
  Leordeanu}.} \bibinfo{year}{2020}\natexlab{}.
\newblock \showarticletitle{Image Difficulty Curriculum for Generative
  Adversarial Networks (CuGAN)}.
\newblock \bibinfo{journal}{\emph{2020 IEEE Winter Conference on Applications
  of Computer Vision (WACV)}} (\bibinfo{year}{2020}),
  \bibinfo{pages}{3452--3461}.
\newblock


\bibitem[\protect\citeauthoryear{Soviany, Ionescu, Rota, and Sebe}{Soviany
  et~al\mbox{.}}{2021}]%
        {Soviany2021CurriculumLearningSurvey}
\bibfield{author}{\bibinfo{person}{Petru Soviany}, \bibinfo{person}{Radu~Tudor
  Ionescu}, \bibinfo{person}{Paolo Rota}, {and} \bibinfo{person}{N. Sebe}.}
  \bibinfo{year}{2021}\natexlab{}.
\newblock \showarticletitle{Curriculum Learning: A Survey}.
\newblock \bibinfo{journal}{\emph{ArXiv}}  \bibinfo{volume}{abs/2101.10382}
  (\bibinfo{year}{2021}).
\newblock


\bibitem[\protect\citeauthoryear{Spitkovsky, Alshawi, and Jurafsky}{Spitkovsky
  et~al\mbox{.}}{2009}]%
        {spitkovsky2009babystepsCL}
\bibfield{author}{\bibinfo{person}{Valentin~I Spitkovsky},
  \bibinfo{person}{Hiyan Alshawi}, {and} \bibinfo{person}{Daniel Jurafsky}.}
  \bibinfo{year}{2009}\natexlab{}.
\newblock \showarticletitle{Baby Steps: How “Less is More” in unsupervised
  dependency parsing}.
\newblock  (\bibinfo{year}{2009}).
\newblock


\bibitem[\protect\citeauthoryear{Suggate, Schaughency, and Reese}{Suggate
  et~al\mbox{.}}{2013}]%
        {suggate2013childlatelearning}
\bibfield{author}{\bibinfo{person}{Sebastian~P Suggate},
  \bibinfo{person}{Elizabeth~A Schaughency}, {and} \bibinfo{person}{Elaine
  Reese}.} \bibinfo{year}{2013}\natexlab{}.
\newblock \showarticletitle{Children learning to read later catch up to
  children reading earlier}.
\newblock \bibinfo{journal}{\emph{Early Childhood Research Quarterly}}
  \bibinfo{volume}{28}, \bibinfo{number}{1} (\bibinfo{year}{2013}),
  \bibinfo{pages}{33--48}.
\newblock


\bibitem[\protect\citeauthoryear{Sutton and Barto}{Sutton and Barto}{2018}]%
        {sutton2018reinforcementlearning}
\bibfield{author}{\bibinfo{person}{Richard~S Sutton} {and}
  \bibinfo{person}{Andrew~G Barto}.} \bibinfo{year}{2018}\natexlab{}.
\newblock \bibinfo{booktitle}{\emph{Reinforcement learning: An introduction}}.
\newblock \bibinfo{publisher}{MIT press}.
\newblock


\bibitem[\protect\citeauthoryear{Sztompka}{Sztompka}{2002}]%
        {sztompka2002socjologia}
\bibfield{author}{\bibinfo{person}{Piotr Sztompka}.}
  \bibinfo{year}{2002}\natexlab{}.
\newblock \showarticletitle{Socjologia}.
\newblock \bibinfo{journal}{\emph{Analiza spo{\l}ecze{\'n}stwa, Znak,
  Krak{\'o}w}} (\bibinfo{year}{2002}), \bibinfo{pages}{324}.
\newblock


\bibitem[\protect\citeauthoryear{Thrun and Pratt}{Thrun and Pratt}{2012}]%
        {thrun2012learningtolearn}
\bibfield{author}{\bibinfo{person}{Sebastian Thrun} {and}
  \bibinfo{person}{Lorien Pratt}.} \bibinfo{year}{2012}\natexlab{}.
\newblock \bibinfo{booktitle}{\emph{Learning to learn}}.
\newblock \bibinfo{publisher}{Springer Science \& Business Media}.
\newblock


\bibitem[\protect\citeauthoryear{Tsutsui, Chandrasekaran, Reza, Crandall, and
  Yu}{Tsutsui et~al\mbox{.}}{2020}]%
        {tsutsui2020computationalToddler}
\bibfield{author}{\bibinfo{person}{Satoshi Tsutsui}, \bibinfo{person}{Arjun
  Chandrasekaran}, \bibinfo{person}{Md~Alimoor Reza}, \bibinfo{person}{David
  Crandall}, {and} \bibinfo{person}{Chen Yu}.} \bibinfo{year}{2020}\natexlab{}.
\newblock \showarticletitle{A Computational Model of Early Word Learning from
  the Infant's Point of View}.
\newblock \bibinfo{journal}{\emph{arXiv preprint arXiv:2006.02802}}
  (\bibinfo{year}{2020}).
\newblock


\bibitem[\protect\citeauthoryear{Tudor~Ionescu, Alexe, Leordeanu, Popescu,
  Papadopoulos, and Ferrari}{Tudor~Ionescu et~al\mbox{.}}{2016}]%
        {tudor2016hardCL}
\bibfield{author}{\bibinfo{person}{Radu Tudor~Ionescu}, \bibinfo{person}{Bogdan
  Alexe}, \bibinfo{person}{Marius Leordeanu}, \bibinfo{person}{Marius Popescu},
  \bibinfo{person}{Dim~P Papadopoulos}, {and} \bibinfo{person}{Vittorio
  Ferrari}.} \bibinfo{year}{2016}\natexlab{}.
\newblock \showarticletitle{How hard can it be? Estimating the difficulty of
  visual search in an image}. In \bibinfo{booktitle}{\emph{Proceedings of the
  IEEE Conference on Computer Vision and Pattern Recognition}}.
  \bibinfo{pages}{2157--2166}.
\newblock


\bibitem[\protect\citeauthoryear{Turchetta, Kolobov, Shah, Krause, and
  Agarwal}{Turchetta et~al\mbox{.}}{2020}]%
        {turchetta2020CISR}
\bibfield{author}{\bibinfo{person}{Matteo Turchetta}, \bibinfo{person}{Andrey
  Kolobov}, \bibinfo{person}{Shital Shah}, \bibinfo{person}{Andreas Krause},
  {and} \bibinfo{person}{Alekh Agarwal}.} \bibinfo{year}{2020}\natexlab{}.
\newblock \showarticletitle{Safe reinforcement learning via curriculum
  induction}.
\newblock \bibinfo{journal}{\emph{arXiv preprint arXiv:2006.12136}}
  (\bibinfo{year}{2020}).
\newblock


\bibitem[\protect\citeauthoryear{Turing}{Turing}{2009}]%
        {turing2009child}
\bibfield{author}{\bibinfo{person}{Alan~M Turing}.}
  \bibinfo{year}{2009}\natexlab{}.
\newblock \showarticletitle{Computing machinery and intelligence}.
\newblock In \bibinfo{booktitle}{\emph{Parsing the Turing Test}}.
  \bibinfo{publisher}{Springer}, \bibinfo{pages}{23--65}.
\newblock


\bibitem[\protect\citeauthoryear{Weiss, Khoshgoftaar, and Wang}{Weiss
  et~al\mbox{.}}{2016}]%
        {weiss2016transfersurvey}
\bibfield{author}{\bibinfo{person}{Karl Weiss}, \bibinfo{person}{Taghi~M
  Khoshgoftaar}, {and} \bibinfo{person}{DingDing Wang}.}
  \bibinfo{year}{2016}\natexlab{}.
\newblock \showarticletitle{A survey of transfer learning}.
\newblock \bibinfo{journal}{\emph{Journal of Big data}} \bibinfo{volume}{3},
  \bibinfo{number}{1} (\bibinfo{year}{2016}), \bibinfo{pages}{9}.
\newblock


\bibitem[\protect\citeauthoryear{Xia, Zamir, He, Sax, Malik, and Savarese}{Xia
  et~al\mbox{.}}{2018}]%
        {xia2018gibsonenv}
\bibfield{author}{\bibinfo{person}{Fei Xia}, \bibinfo{person}{Amir~R Zamir},
  \bibinfo{person}{Zhiyang He}, \bibinfo{person}{Alexander Sax},
  \bibinfo{person}{Jitendra Malik}, {and} \bibinfo{person}{Silvio Savarese}.}
  \bibinfo{year}{2018}\natexlab{}.
\newblock \showarticletitle{Gibson env: Real-world perception for embodied
  agents}. In \bibinfo{booktitle}{\emph{Proceedings of the IEEE Conference on
  Computer Vision and Pattern Recognition}}. \bibinfo{pages}{9068--9079}.
\newblock


\bibitem[\protect\citeauthoryear{Zhang, David, and Gong}{Zhang
  et~al\mbox{.}}{2017}]%
        {Zhang2017CurriculumDAseg}
\bibfield{author}{\bibinfo{person}{Y. Zhang}, \bibinfo{person}{P. David}, {and}
  \bibinfo{person}{Boqing Gong}.} \bibinfo{year}{2017}\natexlab{}.
\newblock \showarticletitle{Curriculum Domain Adaptation for Semantic
  Segmentation of Urban Scenes}.
\newblock \bibinfo{journal}{\emph{2017 IEEE International Conference on
  Computer Vision (ICCV)}} (\bibinfo{year}{2017}), \bibinfo{pages}{2039--2049}.
\newblock


\bibitem[\protect\citeauthoryear{Zosh, Hopkins, Jensen, Liu, Neale,
  Hirsh-Pasek, Solis, and Whitebread}{Zosh et~al\mbox{.}}{2017}]%
        {zosh2017learnthruplayreview}
\bibfield{author}{\bibinfo{person}{Jennifer~N Zosh}, \bibinfo{person}{Emily~J
  Hopkins}, \bibinfo{person}{Hanne Jensen}, \bibinfo{person}{Claire Liu},
  \bibinfo{person}{Dave Neale}, \bibinfo{person}{Kathy Hirsh-Pasek},
  \bibinfo{person}{S~Lynneth Solis}, {and} \bibinfo{person}{David Whitebread}.}
  \bibinfo{year}{2017}\natexlab{}.
\newblock \bibinfo{booktitle}{\emph{Learning through play: a review of the
  evidence}}.
\newblock


\end{thebibliography}

%%
%% If your work has an appendix, this is the place to put it.

\end{document}